\documentclass[fleqn,10pt]{wlscirep}
\usepackage[utf8]{inputenc}
\usepackage[T1]{fontenc}
\usepackage{subfig}%
\usepackage{graphicx}%
\usepackage{amsmath,amssymb,amsfonts}%
\usepackage{amsthm}%
\usepackage{amssymb}
\usepackage{mathtools}
\usepackage[makeroom]{cancel}
\usepackage{array}
\usepackage[title]{appendix}%
\usepackage{xcolor}%
\usepackage{textcomp}%
\usepackage{verbatim}%
\usepackage{manyfoot}%
\usepackage{booktabs}%
\usepackage{multirow}%
\usepackage{algorithm}%
\usepackage{algorithmicx}%
\usepackage{algpseudocode}%
\usepackage{listings}%
\usepackage{float}

\newtheorem{theorem}{Theorem}
\renewenvironment{proof}{{\bfseries Proof: }}{\qed}

\title{Segmenting Action-Value Functions Over Time-Scales in SARSA via TD($\Delta$)}

\author[1,2,3,4,*]{Mahammad Humayoo}

\affil[1]{Hanshan Normal University, Chaozhou, Guangdong, China 521041}
\affil[2]{CAS Key Laboratory of Network Data Science and Technology, Institute of Computing Technology, CAS, Beijing, China 100190}
\affil[3]{University of Chinese Academy of Sciences, Beijing, China 101408}
\affil[4]{School of Computer Science, Beijing Institute of Technology, Beijing, China 100081}

\keywords{SARSA, Temporal Difference (TD($\Delta$)), Action-Value Functions, On-policy, Long-Horizon, Time-Scale Decomposition}

\begin{abstract}
In numerous episodic reinforcement learning (RL) environments, SARSA-based methodologies are employed to enhance policies aimed at maximizing returns over long horizons. Traditional SARSA algorithms face challenges in achieving an optimal balance between bias and variation, primarily due to their dependence on a single, constant discount factor ($\eta$). This investigation enhances the temporal difference decomposition method, TD($\Delta$), by applying it to the SARSA algorithm, now designated as SARSA($\Delta$). SARSA is a widely used on-policy RL method that enhances action-value functions via temporal difference updates. By splitting the action-value function down into components that are linked to specific discount factors, SARSA($\Delta$) makes learning easier across a range of time scales. This analysis makes learning more effective and ensures consistency, particularly in situations where long-horizon improvement is needed. The results of this research show that the suggested strategy works to lower bias in SARSA's updates and speed up convergence in both deterministic and stochastic settings, even in dense reward Atari environments. Experimental results from a variety of benchmark settings show that the proposed SARSA($\Delta$) outperforms existing TD learning techniques in both tabular and deep RL environments.
\end{abstract}
\begin{document}

\flushbottom
\maketitle
%
%
\thispagestyle{empty}


\section{Introduction}
\label{introduction}
Reinforcement learning (RL) agents encounter significant challenges in the optimization of long-horizon rewards. Traditional methods employing singular discount variables encounter inherent challenges: small $\eta$ results in low-variance but myopic policies, whereas large $\eta$ suffers from excessive variance. Fixed discounts remain constant despite variations in reward densities\cite{romoff2019separating,sutton2018reinforcement,bellemare2016unifying}. Temporal difference (TD) learning methods, such as Q-learning and SARSA, have shown effectiveness in various tasks by enabling agents to estimate action-value functions that predict expected future rewards. Traditionally, these strategies use a discount factor of $0\leq\eta<1$, with values getting closer to $\eta=0$ to put more weight on short-term rewards than long-term ones. This approach shortens the planning horizon and makes learning more stable and efficient\cite{sutton2018reinforcement}. Prokhorov and Wunsch\cite{prokhorov1997adaptive} show that discount factors $\eta<1$ often lead to better results in the early stages of learning. In certain scenarios, such as long-horizon tasks\cite{mnih2013playing,berner2019dota} where long-term planning is crucial—like navigation or decision-making tasks that involve delayed rewards—this approach can introduce bias, complicating the agent's ability to determine the most effective long-horizon policies.\par

For example, SARSA\cite{sutton2018reinforcement} (State-Action-Reward-State-Action) is an on-policy RL algorithm that updates the action-value function Q(s,a) according to the state-action transitions the agent sees. The standard SARSA algorithm works well in environments with shorter time horizons; however, it faces challenges when utilized for tasks with longer horizons. The main reason for the observed phenomenon is the use of a single discount factor $\eta$, which requires a trade-off between bias and variance. This often leads to instability and inefficiency during the learning process. A variety of recommendations exist for addressing this issue \cite{berner2019dota,prokhorov1997adaptive,franccois2015discount,xu2018meta,van2009theoretical,romoff2019separating,fedus2019hyperbolic}.\par

Through the use of weighted combinations of value functions, Fedus et al. \cite{fedus2019hyperbolic} presented the concept of hyperbolic discounting. However, their formulation is restricted to state-value functions and does not extend to action-value functions. Meta-gradient RL\cite{xu2018meta} is a technique that enables agents to adapt their learning processes dynamically. Meta-gradient means finding or improving targets and hyperparameters during training, which could make learning more efficient and adaptable. A gradient-based meta-learning algorithm capable of online adaptation to the true value function while engaging with and learning from the environment. However, akin to other methodologies, meta-gradient RL primarily emphasizes state-value functions, rather than action-value formulations.\par

The work that is most relevant in this case is by Romoff et al.\cite{romoff2019separating}, which provides a theoretical basis for temporal difference decomposition using the TD($\Delta$) framework. TD($\Delta$) segments the state-value function V(s) into components using different discount factors, also known as TD($\Delta$). This lets the agent learn about short-term and long-term returns separately, which can then be combined to improve performance. This decomposition framework demonstrates considerable potential in addressing the bias-variance trade-off that is intrinsic to temporal difference learning. However, TD($\Delta$) is constrained to state-value functions, which presents a limitation in the application of multi-timescale representations to action-value-based algorithms such as SARSA and Q-learning.\par

This article proposes SARSA($\Delta$), an enhancement of the TD($\Delta$) approach integrated into the SARSA algorithm, based on this principle. TD($\Delta$) emphasizes the actor-critic framework and conventional TD learning, whereas SARSA($\Delta$) divides the state-action value function into multiple partial estimators corresponding to various discount factors, known as delta estimators. These estimators give a rough estimate of the difference \( D_{m}(s, a) = Q_{\eta_{m}}(s, a) - Q_{\eta_{m -1}}(s, a) \) between action-value functions. This makes learning faster in contexts where actions yield long-term effects. Like TD($\Delta$), SARSA($\Delta$) aims to enhance a collection of delta estimators, with each estimator linked to a specific discount factor. Smaller discount factors lead to quicker convergence via segmentation, whereas larger discount factors enhance this foundation, facilitating improved long-term planning.\par

This paper provides a multi-stage SARSA($\Delta$) updating algorithm that separates action-value functions, which allows for faster convergence and improved policy efficacy, especially in complex deterministic and stochastic environments like Atari games. Moreover, our research shows that SARSA ($\Delta$) can seamlessly accommodate various multi-step RL methodologies, such as n-step SARSA and eligibility traces. This study emphasizes the benefits of SARSA($\Delta$) in various tasks, including a basic ring MDP used by Kearns \& Singh\cite{kearns2000bias}. It outperforms conventional SARSA in scenarios requiring a balance between short-term action consequences and long-term reward maximization. SARSA ($\Delta$) may adjust discount factors for multiple time scales, making it valuable where time scale is precious.\par

The reason for extending TD($\Delta$) to SARSA($\Delta$) is that temporal difference (TD($\Delta$)) learning provides a way to break down value functions across a range of time scales, which solves the problems that come up when learning on a single scale for long-horizon scenarios. We want to break down the action-value function Q(s,a) into components that focus on different discount factors by extending TD($\Delta$) to SARSA($\Delta$). This decomposition enables SARSA to leverage the advantages of learning over multiple time scales, which stabilizes the learning process and reduces the bias-variance trade-off. This add-on gives you a more detailed and scalable way to learn the action-value function in complex environments. Table \ref{Features-table} presents a comparison of the key features of TD($\Delta$)/SARSA($\Delta$) and Standard TD/SARSA-Learning.

\begin{table}[!htbp]
\centering
\caption{Comparison of the essential properties of TD($\Delta$) and SARSA($\Delta$).}
\label{Features-table}
\resizebox{1.0\columnwidth}{!}{   
\begin{tabular}{{|c|c|c|c|c|}}
\hline
\multicolumn{1}{|l|} {Feature} & TD($\Delta$)\cite{romoff2019separating} & SARSA($\Delta$)-Learning (proposed by us) & Standard TD\cite{sutton1998introduction} & SARSA-Learning\cite{sutton1998introduction}\\
\hline
\multicolumn{1}{|l|} {Value function decomposition} & Yes (by $\eta$) & No & No & No \\
\hline
\multicolumn{1}{|l|} {Action-value function decomposition} & No & Yes (by $\eta$) & No & No \\
\hline
\multicolumn{1}{|l|} {Handles multiple time-scales} & Yes & Yes & No & No \\
\hline
\multicolumn{1}{|l|} {Learning stability} & Enhanced, particularly long-term & Enhanced, particularly long-term & Reduced for long-term rewards & Reduced for long-term rewards \\
\hline
\multicolumn{1}{|l|} {Scalability} & Improved & Improved & Restricted by variance/bias & Restricted by variance/bias \\
\hline
\end{tabular}
}
\end{table}
Recent research on value decomposition\cite{romoff2019separating} has demonstrated potential for state-value functions; however, it identifies three significant shortcomings that this work aims to address: (i) Action-Value Formulation: No current method effectively decomposes Q(s, a) while maintaining on-policy guarantees. This study extends TD($\Delta$) to SARSA($\Delta$) by decomposing the action-value function Q(s,a) into delta components across multiple time scales. (ii) Theoretical Foundations: Current analyses do not address the bias-variance trade-off concerning action values. (iii) The proposed SARSA($\Delta$) algorithm improves learning efficiency and stability through the independent learning of components corresponding to different discount factors, which are subsequently integrated to form the comprehensive action-value function. (iv) Both theoretical and empirical evaluations have shown that this multi-scale breakdown has a number of benefits, particularly in environments where there are long-horizon rewards, as demonstrated in Table \ref{Features-table}.\par

The subsequent portions of the paper are organized as follows: Section \ref{relatedwork} presents a discussion on relevant literature. Section \ref{Background} provides the essential background information. The principles of SARSA($\Delta$) and theoretical analysis are shown in Sections \ref{method} and \ref{TA}, respectively. Section \ref{experiment} provides a detailed summary of the experiments performed. Ultimately, we present a conclusion in section \ref{conclusion}.

\section{Related Work}
\label{relatedwork}
Optimizing for long-horizon rewards in reinforcement learning (RL) is difficult owing to the complexities of learning with undiscounted returns. Temporal discounting is often used to simplify this process; however, it can introduce bias.
\subsection{Temporal Decomposition Approaches}
In their investigation, Romoff et al. \cite{romoff2019separating} explored the differentiation of value functions across multiple time scales by scrutinizing individuals with diminished discount factors. The system became more efficient and simpler to expand. In RL, Sherstan et al.\cite{sherstan2018generalizing} introduced $\gamma$-nets, a method designed to enhance value function estimates across various timescales. Authors' recent work \cite{fedus2019hyperbolic,alihyperbolic} showed how to add hyperbolic discounting to RL. Exponential discounting was used in traditional RL, but this didn't match the hyperbolic discounting seen in the behavior of human and animals. The authors showed an agent that used temporal-difference learning methods to get close to hyperbolic discount functions. The research also found that learning value functions across multiple time horizons increases performance, particularly when used as an auxiliary task in value-based RL algorithms like Rainbow. This approach has shown potential in risky and uncertain environments.\par

Recent research \cite{kim2022adaptive,prokhorov1997adaptive,franccois2015discount,xu2018meta,berner2019dota,amit2020discount,wang2020dynamic} focuses on the exact selection of the discount factor. Xu 124
et al. \cite{xu2018meta} proposed meta-gradient methods for choosing discount factors in RL that change over time. They stressed the importance of balancing short- and long-term rewards, which is similar to why we break down action-value functions across multiple discount factors in SARSA($\Delta$). Lastly, hierarchical RL and the bias-variance trade-off are part of a large body of research that is related to our study in an indirect way. These studies \cite{Dietterich1999HierarchicalRL,Henderson2017OptionGANLJ,Hengst2002DiscoveringHI,Reynolds1999DecisionBP,Menache2002QCutD,Russell2003QDecompositionFR,Seijen2017HybridRA} introduced the idea of hierarchical RL breaking value functions down into smaller tasks, similar to how TD($\Delta$) broke value functions down across different time scales. These results backed up the claim that breaking down value functions can make learning more effective and stable.
 \subsection{Multi-Scale RL}
To address issues in RL, especially when the optimum value function is complex and poorly represented by conventional DRL methods,  van Seijen et al. \cite{Seijen2017HybridRA} introduced Hybrid Reward Architectures (HRA). The HRA was designed to enhance learning consistency and effectiveness by partitioning the reward function and allocating distinct value functions to each component. Efficient multi-horizon learning in off-policy RL aims to improve agents' ability to make predictions and decisions over a wide range of time scales. This is important for navigating through complex and unpredictable environments. Ali 141
et al. \cite{ali2022efficient,ali2023multi} investigated designs that enable agents to learn value estimates over several horizons at the same time. This improves their capacity to adjust and flourish in novel environments. In the realm of artificial intelligence and robotics, coordinating agents with limited communication and across diverse planning periods is a significant challenge to multi-horizon, multi-agent planning.\par

Seiler et al. \cite{seiler2024multi} introduced Decentralized Monte Carlo Tree Search (Dec-MCTS) and its variants, which have arisen as formidable instruments for addressing these challenges, allowing agents to plan effectively and adaptively in dynamic, unpredictable environments. Benechehab et al.\cite{benechehab2023multi} came up with a novel technique in model-based RL (MBRL) called multi-timestep models, aimed at mitigating the problem of escalating prediction errors over extended simulated trajectories. By optimizing for several future steps instead of only one-step-ahead forecasts, these models enhance the precision and resilience of long-term predictions, especially in complex or chaotic environments.\par

Bonnet et al. \cite{bonnet2021one} presented multi-step meta-gradient RL, which uses meta-gradients that have been built up across many stages instead of just one. This makes learning algorithms more flexible. This method may provide richer and more useful learning signals, but it does come with some significant trade-offs. But it also increases computational complexity and variation, which may hinder performance if not handled carefully. While deep RL (DRL) can teach agents difficult tasks, it still struggles to improve sampling efficiency and adapt to changing environments. Eligibility traces, a well-established RL approach, expedite learning, but parameter dependencies make their integration with deep neural networks difficult. Kobayashi\cite{kobayashi2022adaptive} introduced several time-scale eligibility traces to equilibrate short-term and long-term credit assignment. This technique improves learning speed and policy quality in online DRL environments by substituting the most important adaptively gathered traces.
 \subsection{SARSA Variants and Traditional RL Methods}
Expected SARSA \cite{van2009theoretical} is a kind of SARSA that speeds up learning by exploiting information from a stochastic behavior policy to create updates that are less variance. Many RL algorithms, like SARSA, employ temporal-difference (TD) learning, which looks at the differences between two predictions made one after the other instead of the final result. Sutton\cite{sutton1988learning} came up with this notion for the first time in 1988. The research shows that TD approaches may be useful for predicting jobs, especially where incremental learning is crucial, and that they can even operate jointly in certain cases.

Theoretical convergence for function approximation-based TD learning algorithms was shown by Tsitsiklis and Van Roy\cite{tsitsiklis1996analysis}. If you follow the SARSA update rule for delta component updates, this research may be used with SARSA($\Delta$) to get the same convergence properties. None of the previously mentioned research discussed short-term estimations to train long-term action-value functions. One benefit is that you may ask about shorter time periods and utilize our method as a generic action-value function. Separating action-value functions across time scales using TD($\Delta$) in SARSA makes a big difference in performance. Theoretical work by  Kearns and Singh\cite{kearns2000bias} helped us comprehend the bias-variance trade-off that comes up when we break down Q-values over multiple time scales. It also set limits on the bias and variance of TD updates. 

Most of the methods listed above focus on state values, although action-value decomposition has not yet been looked at. Our proposed strategy focuses on breaking down $Q(s,a)$ over different time scales, which makes long-term learning more stable.

\section{Background and notation}
\label{Background}
Examine a completely observable Markov Decision Process (MDP) \cite{bellman1957markovian}, characterized by the tuple $(\mathcal{S}, \mathcal{A}, \mathcal{P}, r)$, where $\mathcal{S}$ denotes the state space, $\mathcal{A}$ signifies the action space, and $\mathcal{P}: \mathcal{S} \times \mathcal{A} \rightarrow \mathcal{S} \rightarrow [0, 1] $ represents the transition probabilities that associate state-action pairings with distributions across subsequent states, whereas $r: \mathcal{S} \times \mathcal{A} \rightarrow \mathbb{R}$ denotes the reward function. At each timestep $n$, the agent is state $s_n$, selects an action $a_n$, receives a reward $r_n = r(s_n, a_n)$, and transitions to the next state $s_{n+1} \sim \mathcal{P}(\cdot|s_n, a_n)$.

Within a standard MDP framework, an agent tries to get the highest possible discounted return, which is defined as $Q_{\eta}^{\pi}(s, a) = \mathbb{E}\left[\sum_{n=0}^{\infty} \eta^{n} r_{n+1} \mid s_{n} = s, a_{n} = a\right]$, where $\eta$ is the discount factor and $\pi: \mathcal{S} \rightarrow \mathcal{A} \rightarrow [0, 1]$ stands for the policy that the agent follows. The action-value function $Q_{\eta}^{\pi}(s, a)$ is determined as the fixed point of the Bellman operator $\mathcal{T} Q = r^{\pi} + \eta \mathcal{P}^{\pi} Q$, where $r^{\pi}$ denotes the expected immediate reward and $\mathcal{P}^{\pi}$ represents the transition probability operator associated with the policy $\pi$. For convenience, we remove the superscript $\pi$ for the rest of the paper.\par
Using temporal difference (TD) \cite{sutton1984temporal} learning, the action-value estimate $\hat{Q}_{\eta}$ can approximate the true action-value function $Q_{\eta}$. The one-step TD error $\delta^{\eta}_{n} = r_{n+1} + \eta \hat{Q}_{\eta}(s_{n+1}, a_{n+1}) \mathit{-} \hat{Q}_{\eta}(s_{n}, a_{n})$ is used to update the action-value function given a transition $(s_{n}, a_{n}, r_{n}, s_{n+1})$.

An on-policy RL technique called SARSA learns the action-value function Q(s, a) for a given policy $\pi$. The action-value function represents the expected total reward derived from state s, executing action a, and adhering to policy $\pi$. Then the SARSA update rule is expressed as follows:
\begin{align}
\label{Eq1}
Q(s_{n},a_{n}) \leftarrow Q(s_{n},a_{n}) + \alpha \left[ r_{n} + \eta Q(s_{n+1},a_{n+1}) - Q(s_{n},a_{n})\right]
\end{align}
Here, $\alpha$ is the learning rate, $\eta$ is the discount factor, and $r_{n}$ is the immediate reward. Long-horizon tasks present challenges for SARSA since the choice of $\eta$ can compromise learning efficiency with appropriate long-term reward maximizing.\par
On the other hand, for a complete trajectory, we can employ the discounted sum of one-step TD errors, generally known as the $\lambda$-return \cite{sutton1984temporal} or, equivalently, the Generalized Advantage Estimator (GAE) \cite{Schulman2015HighDimensionalCC}. The GAE improves advantage estimates by balancing the trade-off between variance and bias through the parameters $\lambda$ and $\eta$. The Generalized Advantage Estimator is typically represented by the following equation:
\begin{align}
\label{NGAE}
A(s_{n}, a_{n}) = \sum_{k=0}^{\infty}(\lambda \eta)^{k}\delta_{t+k}^{\eta}
\end{align}
Where $\delta_{n+k}$ is the TD error at time n, computed as follows: $\delta_{n+k} = r_{n+k} + \eta Q(s_{n+k+1}, a_{n+k+1}) - Q(s_{n+k}, a_{n+k})$.\\
Loss function for Q-Value estimation utilizing GAE. The loss function $\mathcal{L}(\theta)$ for approximating the Q-value function is defined as the mean squared error between the current Q-value estimate $Q(s_n, a_n;\theta)$ and the target Q-value adjusted by the advantage estimator. Thus, it is possible to concisely write the loss function using $A(s_n, a_n)$ and $Q(s_n, a_n)$ as follows:
\begin{align}
\label{lossfun}
\mathcal{L}(\theta) = \mathbb{E} \bigg[\bigg(Q(s_n, a_n;\theta) - \bigg(Q(s_n, a_n) + A(s_n, a_n)\bigg)\bigg)^{2}\bigg]
\end{align}
Despite SARSA being on-policy and lacking an explicit policy update, policy selection in SARSA can still be influenced by Eq. \ref{ppoloss}. For actor-critic architectures \cite{Sutton1999PolicyGM,Konda1999ActorCriticA, Mnih2016AsynchronousMF}, the action-value function is updated according to Eq. \ref{lossfun}, and a stochastic parameterized policy (actor, $\pi_\nu(a|s)$) is learned from this value estimator through the advantage function, with the loss being.
\begin{align}
\label{ppoloss}
\mathcal{L}(\nu) = \mathbb{E} \bigg[-log\pi(a|s;\nu)A(s,a)\bigg]
\end{align}
Proximal Policy Optimization (PPO) \cite{schulman2017proximal} is an improvement on actor-critic methods. It limits policy updates to a specific area of optimization called a trust region. It does this by using a clipping objective that compares the current parameters, $\nu$, to the old parameters, $\nu_{old}$:
\begin{align}
\label{ppoobj}
\mathcal{L}(\nu) = \mathbb{E}\bigg[min\bigg(\rho(\nu)A(s,a),\psi(\nu)A(s,a)\bigg)\bigg]
\end{align}
In on-policy approaches such as SARSA, the probability ratio $\rho(\nu)$ between the current policy and a preceding policy is often used. SARSA evaluates actions in accordance with the current policy, concurrently updating the action-value function as actions are performed. However, when utilized within an actor-critic framework, the SARSA-learning agent may adopt a similar policy ratio:
\begin{align}
\label{ppoclipp}
\rho(\nu) = \frac{Q(s, a;\nu)}{Q(s, a;\nu_{old})}
\end{align}
where $\psi(\nu)= clip(\rho, 1-\epsilon, 1+\epsilon)$ represents the clipped likelihood ratio, and $\epsilon < 1$ is a negligible parameter employed to constrain the update. The loss function for the policy update can be articulated as follows:
\begin{align}
\label{cliplossfun}
\mathcal{L}(\nu) = \mathbb{E}\bigg[min\bigg(\rho(\nu)A(s,a),clip(\rho(\nu), 1-\epsilon, 1+\epsilon)A(s,a)\bigg)\bigg]
\end{align}

\section{Methodology}
\label{method}
The TD($\Delta$) paradigm boosts regular TD learning by splitting action-value functions over several discount factors, $\eta_{0}, \eta_{1}, … \eta_{m}$. This breakdown makes it easier to find the action-value function as a sum of delta estimators, where each estimator shows how action-value functions differ at different discount factors. The main advantage of this method is that it makes it easier to control learning variance and bias by focusing on smaller, more controllable time scales.
\subsection{TD($\Delta$) Framework}
\label{TDF}
To compute $D_{m}$ for SARSA as outlined in the study, it is essential to comprehend the mechanism by which delta estimators ($D_{m}$) approximate the discrepancies between action-value functions across successive discount factors. This approach for SARSA will be outlined in a stepwise manner.\\
\textbf{Delta Estimators ($D_{m}$) and Action-Value Functions ($Q(s, a)$)}:
Delta estimators, denoted as $D_{m}$, quantify the variation between action-value functions associated with consecutive discount factors:
\begin{align}
\label{Eq2}
   D_{m} = Q_{\eta_{m}} \mathit{-} Q_{\eta_{m -1}}
\end{align}

In this context, $Q_{\eta_{m}}$ denotes the action-value function with discount factor $\eta_{m}$, but $Q_{\eta_{m-1}}$ signifies the action-value function with the prior discount factor, $\eta_{m-1}$.
\subsection{Single-Step TD ( SARSA($\Delta$) )}
\label{td-sarsa}
This section initiates with an overview of the delta estimators ($D_{m}$) pertaining to the action-value functions $Q$ employed in SARSA. SARSA is expanded to TD($\Delta$), which is called SARSA($\Delta$). SARSA($\Delta$) aims to determine the action-value function $Q(s, a)$, representing the expected return of adhering to policy $\pi$ from state $s$, executing action $a$, and subsequently continuing to follow $\pi$. The goal is to optimize the total reward by improving the action-value function over various time scales. SARSA($\Delta$) fundamentally decomposes the action-value function $Q(s, a)$ into multiple delta components $D_m(s, a)$, each associated with a distinct discount factor $\eta_m$. The relationship between these delta components (i.e., delta function) is expressed in the following manner:
\begin{align} 
\label{destimator1}
 D_{m}(s, a) := Q_{\eta_{m}}(s, a) \mathit{-} Q_{\eta_{m -1}}(s, a) 
 \end{align}
 where $\eta_{0}, \eta_{1}, . . . , \eta_{m}$ represent the discount factors across various time-scales, and define $D_{0}(s,a) := Q_{\eta_{0}}(s, a)$. The action-value function $Q_{\eta_m}(s, a)$ is defined as the cumulative sum of all D-components up to m:
 \begin{align}
 \label{act-val-fun}
 Q_{\eta_m}(s, a) = \sum_{x=0}^m D_x(s, a) 
 \end{align}
The policy dictates how the state-action pair in conventional SARSA determines the update rule. The TD error adjusts the action-value function, and the policy determines the action in the following state:
 \begin{align}
 \label{SARSANorm}
 Q(s_n, a_n) \leftarrow Q(s_n, a_n) + \alpha \left[ r_n + \eta Q(s_{n+1}, a_{n+1}) - Q(s_n, a_n) \right]
 \end{align}
In this context, $r_{n}$ signifies the reward acquired from performing action $a_{n}$ in state $s_{n}$, whereas $(s_{n+1}, a_{n+1})$ denotes the ensuing state-action pair. All of the delta components $D_n(s, a)$ are updated separately using the SARSA rule. The modifications to each component use the same structure as the standard SARSA method, but they are changed to incorporate the delta function at each time scale. The update in Eq. \ref{SARSANorm} for single-step TD SARSA ($\Delta$) can be expressed with numerous time-scales by utilizing the principle of splitting the update into several discount factors \cite{romoff2019separating} as demonstrated below:
 \begin{align}
 \label{SARSADelta} 
 D_m(s_n, a_n) = \mathbb{E} \left[ (\eta_m - \eta_{m-1}) Q_{\eta_{m-1}}(s_{n+1}, a_{n+1}) + \eta_m D_m(s_{n+1}, a_{n+1}) \right]
 \end{align}
In this context, $Q_{\eta_m}(s_n, a_n)$ denotes the action-value function with a discount factor ($\eta_m$). $D_m(s_n, a_n)$ denotes the delta function for SARSA ($\Delta$), while $\eta_m$ represents the discount factor for time-scale m, while $Q_{\eta_m}(s_{n+1}, a_{n+1})$ is the SARSA update for the action-value associated with the subsequent state-action pair. The single-step TD SARSA ($\Delta$) mirrors the original TD update, but it is applied to action-value functions Q rather than state-value functions V. The single-step Bellman Eq. for the Q-value function in standard SARSA is as follows:
 \begin{align}
 \label{BellmanEq}
 Q_{\eta_{m}}(s_{n}, a_{n}) = \mathbb{E}\left[r_{n} + \eta_{n} Q_{\eta_{n}}(s_{n+1}, a_{n+1})\right]
 \end{align}
This denotes the expected value of the reward acquired at time n, in addition to the discounted value of the action-value function in the subsequent state-action pair $(s_{n+1}, a_{n+1})$. Currently, instead of utilizing a single discount factor $\eta$, we are using a series of discount factors $\eta_0, \eta_1, ..., \eta_m$. The delta function $D_m(s_n, a_n)$ is defined as the difference between action-value functions associated with successive discount factors from Eq. \ref{destimator1}:
 \begin{align} 
 \label{destimator2}
 D_m(s_n, a_n) = Q_{\eta_m}(s_n, a_n) - Q_{\eta_{m-1}}(s_n, a_n)
 \end{align}
To expand the delta function $D_m$ utilizing its definition, we replace the Bellman Eq. for both $Q_{\eta_m}$ and $Q_{\eta_{m-1}}$, thereby expressing $D_m$ in relation to these action-value functions. Commence by composing: 
\begin{align} 
\label{singlestep-SARSA-Delta}\nonumber
D_m(s_n, a_n) = Q_{\eta_m}(s_n, a_n) - Q_{\eta_{m-1}}(s_m, a_m) \text{ From Eq. \ref{destimator2}}\\\nonumber
D_m(s_n, a_n) = \mathbb{E}\left[r_n + \eta_m Q_{\eta_m}(s_{n+1}, a_{n+1})\right] - \mathbb{E}\left[r_n + \eta_{m-1} Q_{\eta_{m-1}}(s_{n+1}, a_{n+1})\right] \text{ From Eq. \ref{BellmanEq}}\\\nonumber
\shortintertext{Since the reward terms $r_n$ are included in both formulations, simplify by eliminating them.}
D_m(s_n, a_n) = \mathbb{E}\left[\eta_m Q_{\eta_m}(s_{n+1}, a_{n+1}) - \eta_{m-1} Q_{\eta_{m-1}}(s_{n+1}, a_{n+1})\right] \nonumber
\shortintertext{Breaking down the terms facilitates a more thorough examination of the expression. Utilizing the recursive relationship from Eq. \ref{destimator2}} Q_{\eta_m}(s_{n+1}, a_{n+1}) = D_m(s_{n+1}, a_{n+1}) + Q_{\eta_{m-1}}(s_{n+1}, a_{n+1}) \nonumber
\shortintertext{Substitute the definition of $D_m$ into the Eq. below.}
D_m(s_n, a_n) = \mathbb{E}\left[\eta_m (D_m(s_{n+1}, a_{n+1}) + Q_{\eta_{m-1}}(s_{n+1}, a_{n+1})) - \eta_{n-1} Q_{\eta_{n-1}}(s_{n+1}, a_{n+1})\right] \nonumber
\shortintertext{Make the terms simpler:}
D_m(s_n, a_n) = \mathbb{E}\left[\eta_m D_m(s_{n+1}, a_{n+1}) + (\eta_m - \eta_{m-1}) Q_{\eta_{m-1}}(s_{n+1}, a_{n+1})\right] \nonumber
\shortintertext{Integrate the terms to derive the final update Eq. for $D_m(s_n, a_n)$.}
D_m(s_n, a_n) = \mathbb{E}\left[(\eta_m - \eta_{m-1}) Q_{\eta_{m-1}}(s_{n+1}, a_{n+1}) + \eta_m D_m(s_{n+1}, a_{n+1})\right]
\end{align} 
Eq. \ref{singlestep-SARSA-Delta} shows that the difference in discount factors $\eta_m - \eta_{m-1}$ is integrated into the update for $D_m$ with regard to the action-value function $Q_{\eta_{m-1}}(s_{n+1}, a_{n+1})$ and the bootstrapping from the next step $D_m(s_{n+1}, a_{n+1})$. The Eq. \ref{singlestep-SARSA-Delta} denotes a Bellman Eq. \ref{BellmanEq} for $D_{m}$, combining a decay factor $\eta_{m}$ and the reward $Q_{\eta_{m-1}}(s_{n+1}, a_{n+1})$. Consequently, it can be utilized to define the expected TD update for $D_{m}$. In this expression, $Q_{\eta_{m-1}}(s_{n+1}, a_{n+1})$ can be expressed as the summation of $D_{m}(s_{n+1}, a_{n+1})$ for $m \leq M - 1$, indicating that the Bellman Eq. \ref{BellmanEq} for $D_{m}$ is contingent upon the values of all delta functions $D_{m}$ for $m \leq M- 1$.\\ This approach treats the delta value function at each time-scale as an autonomous RL issue, with rewards obtained from the action-value function of the immediately preceding time-scale. Consequently, for a target discounted action-value function $Q_{\eta_{m}}(s, a)$, all delta components can be trained concurrently through a TD update, employing the prior values of each estimator for bootstrapping. This process necessitates the assumption of a sequence of discount factors, denoted as $\eta_{m}$, which includes both the minimum and maximum values, $\eta_{0}$ and $\eta_{m}$ \cite{romoff2019separating}.
\subsection{Multi-Step TD ( SARSA($\Delta$) )}
\label{mstd-sarsa}
Numerous studies indicate that multi-step TD methods typically exhibit greater efficiency compared to single-step TD methods \cite{sutton1998introduction}. In multi-step TD SARSA($\Delta$), rewards are aggregated over multiple steps instead of depending on a single future reward. This approach takes into account the differences between consecutive discount factors, $\eta_m$ and $\eta_{m-1}$, while utilizing the value function at state $s_{n+k_m}$ for bootstrapping. In SARSA($\Delta$), the update Eqs. are adjusted to combine rewards over a sequence of transitions before bootstrapping from either the Q-values or delta estimators. In multi-step SARSA($\Delta$), the agent accumulates rewards over K steps, resulting in a necessary adjustment to the update rule for each $D_m$.
\begingroup
\allowdisplaybreaks
\begin{align}
\label{MSTDSARSA}
D_m(s_n, a_n) &= \mathbb{E} \biggl[ \sum_{x=1}^{k_m-1} (\eta_m^x - \eta_{m-1}^x) r_{n+x} + (\eta_m^{k_m} - \eta_{m-1}^{k_x}) Q_{\eta_{m-1}}(s_{n+k_m}, a_{n+k_m}) \\ \nonumber
&+ \eta_m^{k_m} D_m(s_{n+k_m}, a_{n+k_m}) \biggr]
\end{align}
\endgroup
In this case, $r_{n+x}$ represents the reward obtained at time step $n+x$. The term $a_{n+k_m}$ and the action are selected in a greedy manner as $a = \arg\max_{a} Q(s_{n+k_m}, a)$, with $k_m$ representing the number of steps associated with the discount factor $\eta_m$. This approach extends Q-learning to multi-step temporal difference learning by gathering rewards over multiple stages and employing bootstrapping from both the present and previous time-scale action-value functions. In normal multi-step TD learning, the TD error is computed by aggregating rewards over several stages rather than relying on a single future step, followed by bootstrapping from the action-value at the last step. For instance, employing a single discount factor $\eta$, the multi-step TD update Eq. can be articulated as:
\begingroup
\allowdisplaybreaks
\begin{align}
Q(s_n, a_n) = \mathbb{E}\left[\sum_{x=0}^{k-1} \eta^x r_{n+x} + \eta^k Q(s_{n+k}, a_{n+k})\right]
\end{align}
\endgroup
Herein, the initial segment of the Eq. aggregates the rewards across k steps, each diminished by $\eta$. The second component derives from the action-value function at time step n+k. We expand this methodology to include various discount factors and delta decomposition, concentrating on representing the delta function $D_m(s_n, a_n)$ as the difference between action-value functions with discount factors $\eta_m$ and $\eta_{m-1}$. According to the definition of $D_m$, we obtain:
\begingroup
\allowdisplaybreaks
\begin{gather}
D_m(s_n, a_n) = Q_{\eta_m}(s_n, a_n) - Q_{\eta_{m-1}}(s_n, a_n)\nonumber
\shortintertext{The multi-step variant of the Bellman Eq. is now implemented for both $Q_{\eta_m}(s_n, a_n)$ and $Q_{\eta_{m-1}}(s_n, a_n)$. The multi-step Bellman Eq. for $Q_{\eta_m}(s_n, a_n)$ is expressed as follows:}
Q_{\eta_m}(s_n, a_n) = \mathbb{E}\biggl[\sum_{x=0}^{k_m - 1} \eta_m^x r_{n+x} + \eta_m^{k_m} Q_{\eta_m}(s_{n+k_m}, a_{n+k_m})\biggr]\nonumber
\shortintertext{For $Q_{\eta_{m-1}}(s_n, a_n)$, the multi-step Bellman Eq. is expressed as follows:}
Q_{\eta_{m-1}}(s_n, a_n) = \mathbb{E}\biggl[\sum_{x=0}^{k_m - 1} \eta_{m-1}^x r_{n+x} + \eta_{m-1}^{k_m} Q_{\eta_{m-1}}(s_{n+k_m}, a_{n+k_m})\biggr]\nonumber
\shortintertext{The formulas for the delta estimator are utilized to perform the subtraction of the two Eqs.}
\begin{split}
D_m(s_n, a_n) &= \biggl[\sum_{x=0}^{k_m - 1} \eta_m^x r_{n+x} + \eta_m^{k_m} Q_{\eta_m}(s_{n+k_m}, a_{n+k_m})\biggr] \\ &- \biggl[\sum_{x=0}^{k_m - 1} \eta_{m-1}^x r_{n+x} + \eta_{m-1}^{k_m} Q_{\eta_{m-1}}(s_{n+k_m}, a_{n+k_m})\biggr]\nonumber
\end{split}
\shortintertext{The expressions are broadened and simplified. The immediate reward terms $r_{n+x}$ are present in both Eqs. but are adjusted by distinct discount factors, $\eta_m$ and $\eta_{m-1}$. This enables us to express the disparity in rewards as:}
\sum_{x=0}^{k_m - 1} \eta_m^x r_{n+x} - \sum_{x=0}^{k_m - 1} \eta_{m-1}^x r_{n+x} = \sum_{x=0}^{k_m - 1} (\eta_m^x - \eta_{m-1}^x) r_{n+x}\nonumber
\shortintertext{Subsequently, for the bootstrapping terms, the recursive relationship is utilized as outlined:}
Q_{\eta_m}(s_{n+k_m}, a_{n+k_m}) = D_m(s_{n+k_m}, a_{n+k_m}) + Q_{\eta_{m-1}}(s_{n+k_m}, a_{n+k_m})\nonumber
\shortintertext{As a result, the following expression is obtained:}
\begin{split}
\eta_m^{k_m} Q_{\eta_m}(s_{n+k_m}, a_{n+k_m}) - \eta_{m-1}^{k_m} Q_{\eta_{m-1}}(s_{n+k_m}, a_{n+k_m}) &= (\eta_m^{k_m} - \eta_{m-1}^{k_m}) Q_{\eta_{m-1}}(s_{n+k_m}, a_{n+k_m}) \\ &+ \eta_m^{k_m} D_m(s_{n+k_m}, a_{n+k_m})\nonumber
\end{split}
\shortintertext{The definitions for both the rewards and bootstrapping are now integrated to yield the final expression:}
\begin{split}
\label{f-multi-TD-SARSA}
D_m(s_n, a_n) &= \mathbb{E}\biggl[\sum_{x=0}^{k_m - 1} (\eta_m^x - \eta_{m-1}^x) r_{n+x} + (\eta_m^{k_m} - \eta_{m-1}^{k_m}) Q_{\eta_{m-1}}(s_{n+k_m}, a_{n+k_m}) \\&+ \eta_m^{k_m} D_m(s_{n+k_m}, a_{n+k_m})\biggr]
\end{split}
\end{gather}
\endgroup
Eq. \ref{f-multi-TD-SARSA} extends the conventional multi-step TD update to include various multiple discount factors and action-value functions in RL. As a result, each $D_{m}$ obtains a share of the rewards from the environment up to time-step $k_{m}-1$. Furthermore, each $D_m$ utilizes its distinct action-value function in conjunction with the value from the prior time scale. A variation of this algorithm, utilizing k-step bootstrapping as described in \cite{sutton1998introduction}, is presented in Algorithm \ref{alg:1}. Despite Algorithm \ref{alg:1} exhibiting quadratic complexity in relation to M, it can be executed with linear complexity for substantial M by preserving $\hat{Q}$ action-values at each time-scale $\eta_{m}$.
\begin{algorithm}[!htbp]
   \caption{Multi-Step TD ( SARSA($\Delta$) )}
   \label{alg:1}
\begin{algorithmic}
   \State  {\bfseries Inputs:} Pick out the discount factors ($\eta_0, \eta_1, ..., \eta_M$), bootstrapping steps ($k_0, k_1, ..., k_M$), and learning rates ($\alpha_0, \alpha_1, ..., \alpha_M$).
   \State Set the initial value of $D_m(s, a) = 0$ for all states, actions, and scales m.
   \For{episode = 0, 1, 2,...}\Comment{Loop for each episode}
   \State Initialize state $s_0$ and choose an initial action $a_0$ in accordance with a policy.
   \For{n = 0, 1, 2,...} \Comment{Loop for each time step}
   \State Take action $a_n$, observe reward $r_n$ and next state $s_{n+1}$.
   \State Select action $a_{n+1}$ based on a policy.
   \For{m = 0, 1, ..., M}
   \If{m = 0}
   \State $G^0 = \sum_{x=0}^{k_0 - 1} \eta_0^x r_{n+x} + \eta_0^{k_0} D_0(s_{n+k_0}, a_{n+k_0})$
   \Else \Comment{Utilizing Eq. \ref{act-val-fun}, we substitute $Q_{\eta_{m-1}}(s_{n+k_m}, a_{n+k_m})$ by summing the D-components up to $D_{m-1}$ in Eq. \ref{f-multi-TD-SARSA}.}
    \State $G^m = \sum_{x=0}^{k_m - 1} (\eta_m^x - \eta_{m-1}^x) r_{n+x} + (\eta_m^{k_m} - \eta_{m-1}^{k_m}) \sum_{m=0}^{m-1} D_{m-1}(s_{n+k_m}, a_{n+k_m}) + \eta_m^{k_m} D_m(s_{n+k_m}, a_{n+k_m})$
   \EndIf

   \EndFor
   \For {m = 0, 1, 2,..., M}
   \State $D_m(s_n, a_n) \leftarrow D_m(s_n, a_n) + \alpha_m (G^m - D_m(s_n, a_n))$
   \EndFor
   \EndFor
   \EndFor
\end{algorithmic}
\end{algorithm}

\subsection{SARSA TD($\lambda, \Delta$)}
\label{SARSA-TD-LAMBDA-Delta}
Eq. \ref{STD-Return} presents the $\lambda$-return \cite{sutton1984temporal,sutton2018reinforcement}, which integrates rewards across several steps to establish a target for TD($\lambda$) updates. The $\lambda$-return $ G_{n}^{\eta,\lambda}$ is formally defined as follows:
\begin{equation}
\label{STD-Return}
   G_{n}^{\eta,\lambda}(s_n,a_n) = \hat{Q}_{\eta}(s_{n}, a_{n})+ \sum_{k=0}^{\infty}(\lambda\eta)^{k}\delta_{n+k}^{\eta}
\end{equation}
Eq. \ref{bellmaneq} defines the TD($\lambda$) operator, which is used to iteratively apply $\lambda$-returns in updating the value functions. For SARSA, the TD($\lambda$) operator updates the action-value function by summing the $\lambda$-discounted TD errors as follows:
\begin{equation}
\label{bellmaneq}
\mathrm{T}_{\lambda}Q(s_n, a_n) = Q(s_n, a_n) + (I - \lambda \eta P)^{-1}( \mathrm{T}Q(s_n, a_n) - Q(s_n, a_n))
\end{equation}
P represents the transition matrix for state-action pairs according to the policy $\pi$, while Q denotes the action-value function.
Similarly, Eq. \ref{SRD-Estimator} defines the $\lambda$-return specific to delta estimators in TD($\lambda, \Delta$), denoted as $G_{n}^{m,\lambda_{m}}$ for each delta estimator $D_{m}$:
\begin{equation}
\label{SRD-Estimator}
   G_{n}^{m,\lambda_{m}} := \hat{D}_{m}(s_{n}, a_{n})+ \sum_{k=0}^{\infty}(\lambda_{m}\eta_{m})^{k}\delta_{n+k}^{m}
\end{equation}
where $\delta_{n}^{0}:=\delta_{n}^{\eta_{0}}$ and $\delta_{n}^{m} := (\eta_{m} - \eta_{m-1} )\hat{Q}_{\eta_{m-1}}(s_{n+1}, a_{n+1}) + \eta_{m} \hat{D}_{m}(s_{n+1},a_{n+1}) \\- \hat{D}_{m}(s_{n+1}, a_{n+1})$ are the TD-errors.
\subsection{SARSA TD($\lambda, \Delta$) with Generalized Advantage Estimation (GAE)}
\label{GAE}
Generalized Advantage Estimation (GAE) \cite{schulman2017proximal} seeks to compute the advantage function by aggregating multi-step TD errors. To apply this approach to delta estimators $D_{m}(s_{n},a_{n})$, advantage estimates $A^{\Delta}(s_{n}, a_{n})$ are computed for each time scale. Each advantage estimate $A^{\Delta}(s_{n}, a_{n})$ employs a multi-step TD error pertinent to the delta estimator $D_n$, expressed as follows:
\begin{equation}
\label{sgae}
   A^{\Delta}(s_{n}, a_{n}) = \sum_{k=0}^{T-1}(\lambda_{m}\eta_{m})^{k}\delta_{n+k}^{\Delta}
\end{equation}
where $\delta_{n+k}^{\Delta} := r_{n} + \eta_{m}\sum_{m=0}^{M}\hat{D}_{m}(s_{n+1},a_{n+1}) - \sum_{m=0}^{M}\hat{D}_{m}(s_{n}, a_{n})$.\\
The discount factor $\eta_{m}$ is utilized, and the sum of all D estimators as a surrogate for $Q_{\eta m}$. This objective is applicable to PPO by implementing the policy update from Eq. \ref{cliplossfun} and replacing A with $A^{\Delta}$. Additionally, to train each $D_{m}$, a truncated form of their corresponding $\lambda$-return are used, as outlined in Eq. \ref{SRD-Estimator}. For more information, see Algorithm \ref{alg:2}.

\begin{algorithm}[!htbp]
   \caption{PPO-TD($\lambda$, SARSA($\Delta$))}
   \label{alg:2}
\begin{algorithmic}
   \State  {\bfseries Inputs:} Pick out the discount factors ($\eta_0, \eta_1, ..., \eta_M$), bootstrapping steps ($k_0, k_1, ..., k_M$), and learning rates ($\alpha_0, \alpha_1, ..., \alpha_M$).
   \State Set the initial value of $D_m(s, a) = 0$ for all states s, actions a, and scales m.
   \State Initialize policy $\nu$, and values $\theta^{m} \forall m$
   \For{episode = 0, 1, 2,...}\Comment{Loop for each episode}
   \State Initialize state $s_0$ and choose an initial action $a_0$ in accordance with a policy.
   \For{n = 0, 1, 2,...} \Comment{Loop for each time step}
   \State Take action $a_n$, observe reward $r_n$ and next state $s_{n+1}$.
   \State Select next action $a_{n+1}$ based on a policy.
   \For{m = 0, 1, ..., M}
   \If{$n \geq T$}
   \State $G^{m, \lambda_m} \leftarrow \hat{D}_m(s_{n-T}, a_{n-T}) + \sum_{k=0}^{T-1} (\lambda_m \eta_m)^k \delta_{n-T+k}^m \forall m$ \Comment{Computing multi-step return $G^{m, \lambda_m}$ and  TD-error $\delta_{n-T+k}^m$ using Eq. \ref{SRD-Estimator}.}
   \EndIf
   \EndFor
   \For {m = 0, 1, 2,..., M}
   \State $\hat{D}_m(s_{n-T}, a_{n-T}) \leftarrow \hat{D}_m(s_{n-T}, a_{n-T}) + \alpha_m \left(G^{m, \lambda_m} - \hat{D}_m(s_{n-T}, a_{n-T})\right)$
   \State $A^{\Delta} = \sum_{k=0}^{T-1} (\lambda_m \eta_m)^k \delta_{n-T+k}^{\Delta}$ \Comment{Where $A^{\Delta}$ and $\delta_{n-T+k}^{\Delta}$ are computed using Eq. \ref{sgae}.}
   \State $\theta^m \leftarrow \theta^m + \alpha_m \left( G^{m, \lambda_m} - \hat{D}_m(s_{n-T}, a_{n-T}) \right) \nabla \hat{D}_m(s_{n-T}, a_{n-T})$ \Comment{Update $\theta^m$ with TD (Eq. \ref{lossfun}) using $G^{m, \lambda_m} \forall m$.}
   \State $\mathcal{L}(\nu) = \mathbb{E}\bigg[min\bigg(\rho(\nu)A^{\Delta}(s,a),clip(\rho(\nu), 1-\epsilon, 1+\epsilon)A^{\Delta}(s,a)\bigg)\bigg]$ \Comment{from Eq. \ref{cliplossfun}}
   \State $\nu \leftarrow \nu + \alpha_\nu \nabla_\nu \mathcal{L}(\nu)$ \Comment{Update the policy parameters $\nu$ with PPO (Eq. \ref{cliplossfun}) for SARSA using $A^{\Delta}$}
   \EndFor
   \EndFor
   \EndFor
\end{algorithmic}
\end{algorithm}

\section{Analysis}
\label{TA}
Subsequently, the delta estimators are analyzed in relation to the bias-variance trade-off. In SARSA, bias arises from reliance on current estimates, influencing updates over short time scales, especially with smaller discount factors, like lower values of $\eta$ in $D_m$. Utilizing smaller discount factors in updates leads to a decrease in variance because they depend on current estimates. However, this approach also results in heightened bias, as it does not incorporate significant information about future rewards. Conversely, variance increases for updates linked to longer time scales, as future reward information introduces additional variability. In TD($\Delta$), action-value components $D_m$ with higher $\eta_m$ are associated with greater long-term rewards; however, this results in increased variance stemming from the stochastic characteristics of reward sequences. We begin by demonstrating that our estimator is equivalent to the standard estimator $\hat{Q}_{\eta}$ under specific conditions, as outlined in Theorem \ref{thm:thm1}. This comparison elucidates the essential metrics of our estimator that may indicate potential advantages over the standard $\hat{Q}_{\eta}$ estimator. Building on this result and previous research by Kearns and Singh \cite{kearns2000bias}, the analyses are adapted for SARSA to investigate the effects of bias and variance, along with the TD($\Delta$) decomposition framework, in relation to action-value functions. Our objective is to extend the bias-variance error bound framework to action-value settings using TD($\Delta$)(i.e., Theorem \ref{thm:thm4}), thereby offering a better understanding of how these quantities can be balanced to obtain optimal results \cite{romoff2019separating}.
\subsection{SARSA Equivalency Configurations and Enhancements}
\label{SECE}
In certain scenarios, we can demonstrate that our delta estimators for SARSA correspond with the traditional action-value estimator when reformulated into an action-value function. This discourse focuses on the approximation of the linear function of the specified form:\\
$\hat{Q}(s, a)_{\eta} := \langle \theta^{\eta}, \phi(s, a) \rangle \quad \text{and} \quad \hat{D}_m(s, a) := \langle \theta^m, \phi(s, a) \rangle, \forall m$\\
where $\theta$ and $\theta^{m}$ represent weight vectors in $\mathbb{R}^d$ and the function $\phi: S \times A \rightarrow \mathbb{R}^d$ represents a mapping from a state-action pair to a specified d-dimensional feature space. The weight vector $\theta$ for SARSA is updated according to the TD($\lambda$) learning rule in the following manner:
\begin{align}
\label{Delta-Gae}
   \theta_{n+1}^{\eta} = \theta_{n}^{\eta} + \alpha \bigg(G_{n}^{\eta, \lambda} - \hat{Q}_{\eta}(s, a)\bigg)\phi(s_{n},a_{n}),
\end{align}
Here, $G_{n}^{\eta,\lambda}$ denotes the TD($\lambda$) return as stated in Eq. \ref{STD-Return}. Likewise, the delta estimator approach TD($\lambda_{m},\Delta$) is used to update each $\hat{D}_{m}$:
\begin{align}
\label{Deltaestimator}
   \theta_{n+1}^{m} = \theta_{n}^{m} + \alpha \bigg(G_{n}^{M, \lambda_{m}} - \hat{D}_{m}(s_{n}, a_{n})\bigg)\phi(s_{n},a_{n}),
\end{align}
where $G_{n}^{M, \lambda_{m}}$ is specified identically to the TD($\Delta$) return defined in Eq. \ref{SRD-Estimator}, modified for the particular action-value function. In these equations, $\alpha$ and $\{\alpha_m\}_m$ represent positive learning rates. Two SARSA version algorithms are shown to be equivalent in this context by the following theorem. The subsequent theorems resemble those presented by  Romoff et al.\cite{romoff2019separating}; however,  a proof is presented for the action-value function in relation to SARSA, whereas Romoff et al. established the proof for the value function.
\begin{theorem}
\label{thm:thm1}
Inspired by Romoff et al.\cite{romoff2019separating}, if $\alpha_{m} = \alpha$, $\lambda_{m}\eta_{m} = \lambda_{\eta}$, $\forall m$, and if we choose the initial conditions in such a way that $\sum_{m=0}^{M}\theta_{0}^{\eta} = \theta_{0}^{\eta}$, then the iterates produced by TD($\lambda$) (Eq. \ref{Delta-Gae}) and TD($\lambda, \Delta$) (Eq. \ref{Deltaestimator}) with linear function approximation satisfy the following conditions:
\begin{align}
\label{thm:eq1}
\Sigma_{m=0}^{M}\theta_{n}^{m} = \theta_{n}^{\eta}
\end{align}
\end{theorem}
The proof is presented in appendix \ref{proofs}.\\
Equivalence is attained when $\lambda_{m}\eta_{m}=\lambda_{\eta},\forall m$. When $\lambda$ approaches 1 and $\eta_{m}$ is less than $\eta$, this condition suggests that $\lambda_{m}=\lambda_{\eta}/\eta_{m}$ could potentially surpass one, resulting in a risk of divergence in TD($\lambda_{m}$). The subsequent theorem demonstrates that the TD($\lambda$) operator, as defined in Eq. \ref{bellmaneq}, qualifies as a contraction mapping for the range $1 \leq \lambda < \frac{1+\eta}{2\eta}$, thereby confirming that $\lambda_{\eta}<1$ \cite{romoff2019separating}.

\begin{theorem}
\label{thm:thm2}
Inspired by Romoff et al.\cite{romoff2019separating}, $\forall\lambda \in \left[ 0, \frac{1 + \eta}{2\eta} \right]$, the operator $T_{\lambda}$ is defined by the Eq.  $T_{\lambda} Q = Q + (I - \lambda \eta P)^{-1} (T Q - Q)$ $,\quad \forall Q \in \mathbb{R}^{|S| \times |A|}$, is well-defined. Furthermore, $T_{\lambda} Q$ constitutes a contraction in relation to the max norm, with its contraction coefficient expressed as $\frac{\eta}{|1 - \lambda \eta|}$.
\end{theorem}
The proof is presented in appendix \ref{proofs}.\\
The analysis presented in Theorem \ref{thm:thm1} can be adapted to a new context in which $k_m$-step TD ($\Delta$) is employed for each $D_m$ component, rather than TD($\lambda$, $\Delta$), a variant of TD learning. Theorem \ref{thm:thm1} demonstrates that with linear function approximation, standard multi-step TD and multi-step TD($\Delta$) can be equivalent when the number of steps ($k_m$) is consistent across all time scales (i.e., $k_m = k, \forall m$).\\
Although these methods could theoretically be equivalent, such equivalence is not common. To maintain the preservation of equivalence, it is crucial for the learning rate to remain consistent across all time scales. This underscores a significant limitation, as shorter time scales, which involve fewer future steps, can be acquired more rapidly than longer time scales that require the assessment of additional steps beforehand. Furthermore, in practical applications, particularly involving nonlinear function approximation such as deep neural networks, adaptive optimizers are frequently employed \cite{henderson2018did, schulman2017proximal}. The optimizers modify the learning rate accordance with the complexity of the learning task, which depends on the attributes of the delta estimator and its target. There is no universally applicable learning rate in this setting since the effective learning rate changes with time scale.\\
In addition to the learning rate, breaking the action-value function into separate components (especially the $D_m$ components) has advantages that standard TD learning, such as non-delta estimators, does not have.  The ability to utilize various k-step returns (or $\lambda$-return) across multiple time scales enhances control and adaptability in the learning process. If $k_m < k_{m+1},\forall m\quad(or \quad \eta_{m}\lambda_{m} < \eta_{m+1}\lambda_{m+1} , \forall m)$, this approach has the potential to reduce variance while simultaneously introducing bias, since shorter time scales, which can be learned more quickly, may not correspond perfectly with longer time scales regarding the values they estimate.
\subsection{Evaluation for Minimizing $k_{m}$ Values in SARSA via Phased Updates}
\label{arv}
To show how our approach differs from the single estimator case, let's follow the tabular phased version of k-step TD presented by Kearns and Singh\cite{kearns2000bias}. In SARSA, the goal is to evaluate the value associated with a state-action pair (s, a) starting from each state $s \in \mathcal{S}$ and with action $a \in \mathcal{A}$, then n trajectories are generated $\bigg\{ S_{0}^{(x)} = s, a_{0}, \eta_{0}, ..., S_{k}^{(x)}, a_{k}^{(x)}, \eta_{k}^{(x)}, S_{k+1}^{(x)}, ...\bigg\}_{1\leq x \leq n}$ by following the policy $\pi$ and averaging over trajectories. Let $Q_{\eta, n}(s, a)$ denote the Q-value estimate at phase n for the discount factor $\eta$. For each iteration n, also called phase n, obtain the phase-based estimate $Q_{\eta, n}(s, a)$ for (s, a); we average over the trajectories, resulting in:
\begin{align}
\hat{Q}_{\eta, n}(s, a) =\frac{1}{n} \sum_{x=1}^{n}\bigg( \sum_{i=0}^{k-1} \gamma^{i}r_{i}^{(x)} + \eta^{k} \hat{Q}_{\eta,n-1}(s_{k}^{(x)}, a_{k}^{(x)})\bigg)
\end{align}
Theorem \ref{thm:thm3} draws upon the findings of Kearns and Singh\cite{kearns2000bias}, who demonstrated the result concerning the state-value function. In contrast, the subsequent theorem introduces an upper limit on the error associated with action-value function estimation, expressed as $\Delta_{n}^{\hat{Q}_{\eta}} = max_{s, a}\bigg\{\bigg|\hat{Q}_{\eta, n}(s, a) - Q_{\eta}(s, a)\bigg|\bigg\}$

\begin{theorem}
\label{thm:thm3}
Inspired by Kearns and Singh\cite{kearns2000bias}, for any $0<\delta<1$, let $\epsilon = \sqrt{\frac{2log(2k/\delta)}{n}}$. with probability $1-\delta$,
\begin{align}
\label{thm3:Eq}
\Delta_{n}^{\hat{Q}_{\eta}} \leq \epsilon \bigg(\underbrace{\frac{1-\eta^{k}}{1-\eta} \bigg)}_{\text{variance term}}+ \underbrace{\eta^{k} \Delta_{n-1}^{\hat{Q}_{\eta}}}_{\text{bias term}}
\end{align}
\end{theorem}
The proof is presented in appendix \ref{proofs}.\\

In Eq. \ref{thm3:Eq}, a variance term ($\frac{1-\eta^{k}}{1-\eta}$) is arising  due to sampling error from rewards collected along trajectories. In particular, $\epsilon$ bounds the deviation of the empirical average of rewards from the true expected reward. The second term ($\eta^{k} \Delta_{n-1}^{\hat{Q}_{\eta}}$) in Eq. \ref{thm3:Eq} is a bias term that arises from bootstrapping off the previous phase's estimates.

Similarly, a phased SARSA variant of the multi-step $TD(\Delta)$ method is considered. For each phase n, each D component is updated as follows:
\begin{align}
\hat{D}_{m,n}(s,a) = \frac{1}{n}\sum_{x=1}^{n}\biggl(\sum_{i=1}^{k-1}(\eta_{m}^{i} - \eta_{m-1}^{i})r_{i}^{(x)} + (\eta_{m}^{k_{m}} - \eta_{m-1}^{k_{m}})Q_{\eta_{m-1}}(s_{n+k}^{(x)}, a_{n+k}^{(x)}) + \eta_{m}^{k_{m}} \hat{D}_{m}(s_{n+k}^{(x)}, a_{n+k}^{(x)})\biggr)
\end{align}
The maximum threshold for the phased error has now been established. The cumulative errors of each D component, denoted as $TD(\Delta)$, is expressed as $\sum_{m=0}^{M}\Delta_{n}^{m}$, where $\Delta_{n}^{m}= \max_{s, a}\bigg\{\bigg | \hat{D}_{m}(s,a) - D_{m}(s,a) \bigg |\bigg\}$. Essential differences between SARSA and Q-learning regarding the error bound: In SARSA, the error is defined by the actions taken by the agent, indicating that the Q-values are modified based on the state-action pairings encountered by the agent during exploration. The inaccuracy in Q-learning emerges from the greedy action, which is defined as the action that maximizes the Q-value for the future state, assuming that the agent consistently operates in an optimal manner.

\begin{theorem}
\label{thm:thm4}
Inspired by Romoff et al.\cite{romoff2019separating}, assume that $\eta_{0} \leq \eta_{1} \leq \eta_{2}\leq ..... \leq \eta_{m}=\eta$ and $K_{0} \leq K_{1} \leq ... \leq K_{m}= K$, for any $0 < \delta < 1$, let $\epsilon = \sqrt{\frac{2log(2k/\delta)}{n}}$, with probability $1 - \delta$,
\begin{align}
\label{thm4:eq1}
\sum_{m=0}^{M} \Delta_{n}^{m} \leq \epsilon \bigg( \frac{1-\eta^{k}}{1 - \eta} \bigg)+ \epsilon \bigg(\underbrace{\sum_{m=0}^{M-1} \frac{\eta_{m}^{k_{m+1}}-\eta^{k_{m}}_{m}}{1-\eta_{m}} \bigg)}_{\text{variance reduction}}+ \underbrace{\sum_{m=0}^{M-1}\bigg( \eta^{k_{m}}_{m} - \eta^{k_{m+1}}_{m} \bigg)\sum_{q=0}^{m}\Delta_{n-1}^{q}}_{\text{bias introduction}} + \eta^{k}\sum_{m=0}^{M}\Delta_{n-1}^{m}
\end{align}
\end{theorem}
The proof is presented in appendix \ref{proofs}.\\
The authors examine the comparison of the limits for phased TD($\lambda$) in theorem \ref{thm:thm3} with those for phased TD($\Delta$) in theorem \ref{thm:thm4}. This comparison demonstrates that phased TD($\Delta$) facilitates variance reduction equivalent to $\epsilon\sum_{m=0}^{M-1} \frac{\eta_{m}^{k_{m+1}}-\eta^{k_{m}}_{m}}{1-\eta_{m}}\leq 0$ but introduces a potential bias quantified as $\sum_{m=0}^{M-1}\bigg( \eta^{k_{m}}_{m} - \eta^{k_{m+1}}_{m} \bigg)\sum_{q=0}^{m}\Delta_{n-1}^{q} \geq 0$. Utilizing phased TD($\Delta$) to decrease $k_{m}$ synchronizes updates more effectively with recent actions, hence lowering variance from high-discounted return components. Shortened $k_{m}$ values accrue bias more swiftly, as SARSA bootstraps from current policy estimations that may vary over phases. It is observed that when all $k_{m}$ values are identical, both algorithms yield the same upper bound.

In SARSA, the return estimate includes subsequent states and actions determined by the policy. The anticipated discounted return over T steps closely approximates the infinite-horizon discounted expected return after $\mathrm{T}$, where $\mathrm{T}\approx\frac{1}{1-\eta}$ \cite{kearns2002near}. Consequently, $k_{m}$ can be effectively simplified for any $\eta_{m}$ such that $k_{m}\approx \frac{1}{1-\eta_{m}}$, thereby adhering to this principle. Therefore, with $\mathrm{T}$ samples, similar to TD($\Delta$) \cite{romoff2019separating}, SARSA employing TD($\Delta$) utilizes $k_{m}\approx \frac{1}{1-\eta_{m}}$ to establish a suitable balance between bias and variance over all time scales significantly less than $\mathrm{T}$. By establishing $k_{m}$ for each $\eta_{m}$, where $\eta_{m}^{\frac{1}{1-\eta_{m}}}\leq \frac{1}{e}$, the approach guarantees that the effective horizon and variance are maintained within limits without requiring extensive parameter optimization. Each increase in $D_m$ facilitates the doubling of the effective horizon, hence ensuring logarithmic scaling with the quantity of action-value functions. SARSA can utilize this to modify additional parameters in alignment with the specified time scales, enhancing the algorithm's efficiency in calculating long-term rewards without added complexity.

\section{Experiments and Results}
\label{experiment}
\begin{figure}[!htbp]
\centering
\subfloat[FrozenLake-v1.]{
\includegraphics[width=0.20\columnwidth, height=0.20\columnwidth]{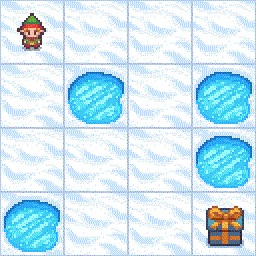}
\label{envlake}
}%
\subfloat[Taxi-v3.]{
\includegraphics[width=0.20\columnwidth, height=0.20\columnwidth]{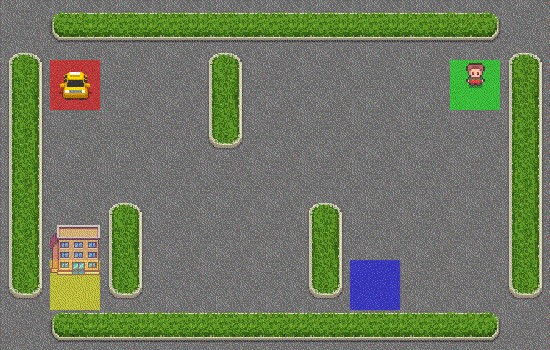}
\label{envtaxi}
}%
\subfloat[Blackjack-v1.]{
\includegraphics[width=0.20\columnwidth, height=0.20\columnwidth]{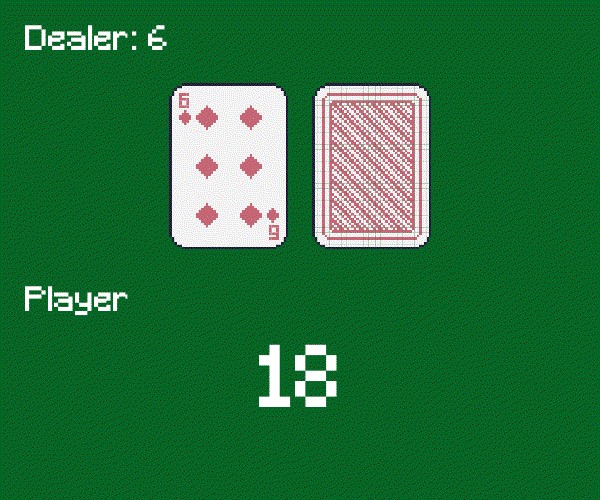}
\label{envblackjack}
}
\subfloat[CliffWalking-v0.]{
\includegraphics[width=0.20\columnwidth, height=0.20\columnwidth]{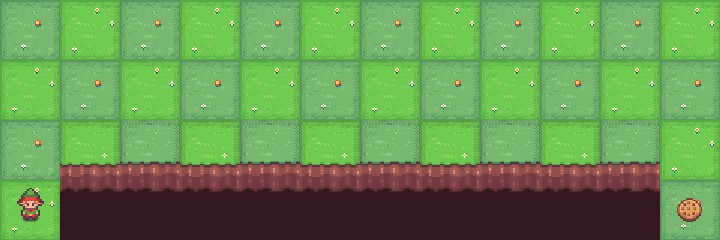}
\label{envlunar}
}%
\caption{OpenAI Gym control used for all experiments. In order from left to right: Frozen Lake, Taxi, Blackjack, and Cliff Walking.}
\label{allenv}
\end{figure}

\begin{figure}[!htbp]
\centering
\subfloat[SeaquestNoFrameskip-v4]{
\includegraphics[width=0.40\columnwidth, height=0.40\columnwidth]{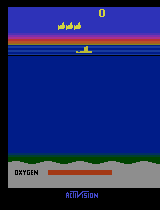}
\label{seaquest}
}%
\subfloat[QbertNoFrameskip-v4]{
\includegraphics[width=0.40\columnwidth, height=0.40\columnwidth]{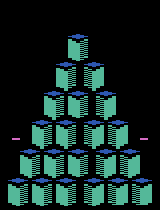}
\label{qbert}
}%
\caption{The OpenAI Gym Atari control framework is utilized for conducting long-horizon experiments. Arranged from left to right: \textbf{(a)} SeaquestNoFrameskip-v4 and \textbf{(b)} QbertNoFrameskip-v4.}
\label{fig:qbert4-seaquest4}
\end{figure}

Experiments were conducted in three distinct environments to evaluate the effectiveness of the proposed method: the Tabular Ring MDP environment detailed in Section \ref{tabular}, the deterministic and stochastic OpenAI Gym environments examined in Section \ref{SecOpenAI}, and the Atari environments from OpenAI Gym described in Section \ref{secAtari}. This section initiates with a detailed overview of the experimental setup.\\
\textbf{Experimental Setup:} Experiments were conducted using OpenAI Gym control tasks \cite{brockman2016openai}, and Figs.\ref{rmdp}, \ref{allenv}, and \ref{fig:qbert4-seaquest4} show the relevant environments. The experiments were carried out on a single PC with 16 GB of memory, an Intel Core i7-2600 processor, and a GPU. The machine ran on 64-bit Ubuntu 22.04.5 LTS. Python 3.9 was used during development, along with the PyTorch package. In each experiment, ten randomly selected seeds were utilized, and the average results are provided. The remaining hyperparameter settings are displayed in the table \ref{hyperparameter-table} of appendix \ref{appendix-hyper}.
\subsection{Tabular}
\label{tabular}
Q-value iteration is utilized to determine the optimal Q-value function. An open-source implementation of Value Iteration (VI), based on the code by Peter Henderson\cite{value-iteration}, is employed in accordance with the algorithmic principles outlined in Bellman\cite{bellman1957markovian} and Sutton et al.\cite{sutton1998introduction}. The methodology calculates the optimal value function; however, this approach is adjusted to ascertain the optimal Q-value function. Upon acquiring the optimal Q-values, Temporal Difference (TD) and SARSA($\Delta$) methods are applied to derive approximation Q-values via sampling. After 5000 steps, each learning iteration comes to an end. For every setup, 200 trials are executed using different random seeds that influence the dynamics of the transition. The algorithms are implemented using the codebase provided by Kostrikov\cite{pytorchrl}.

\begin{figure}[!htbp]
\centering
\includegraphics[width=0.5\columnwidth]{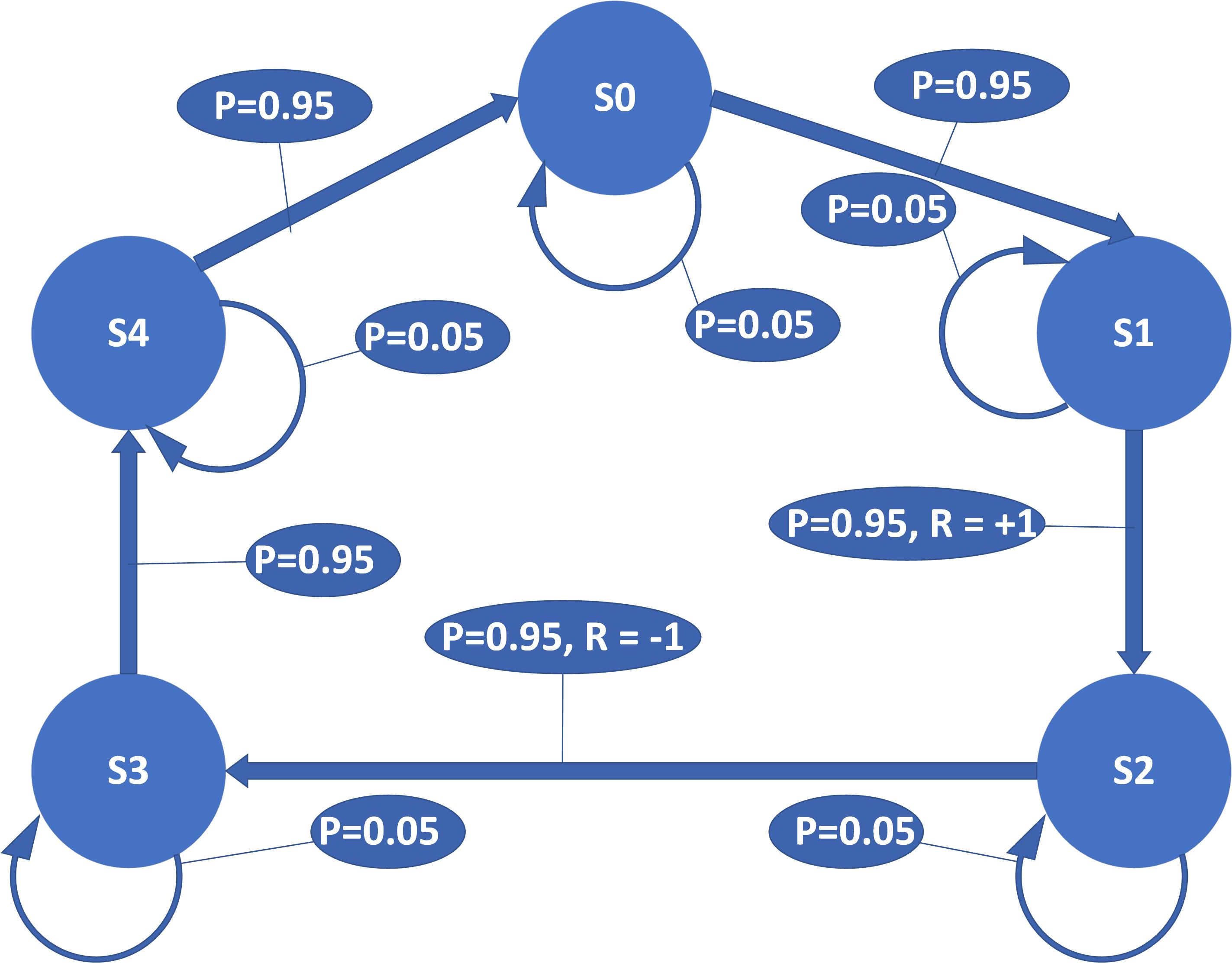}
\caption{Five-state ring MDP as described by Kearns and Singh\cite{kearns2000bias}}\label{rmdp}.
\end{figure}

\begin{figure}[!htbp]
\centering
\subfloat[Error at different learning rates.]{
\includegraphics[width=0.5\columnwidth]{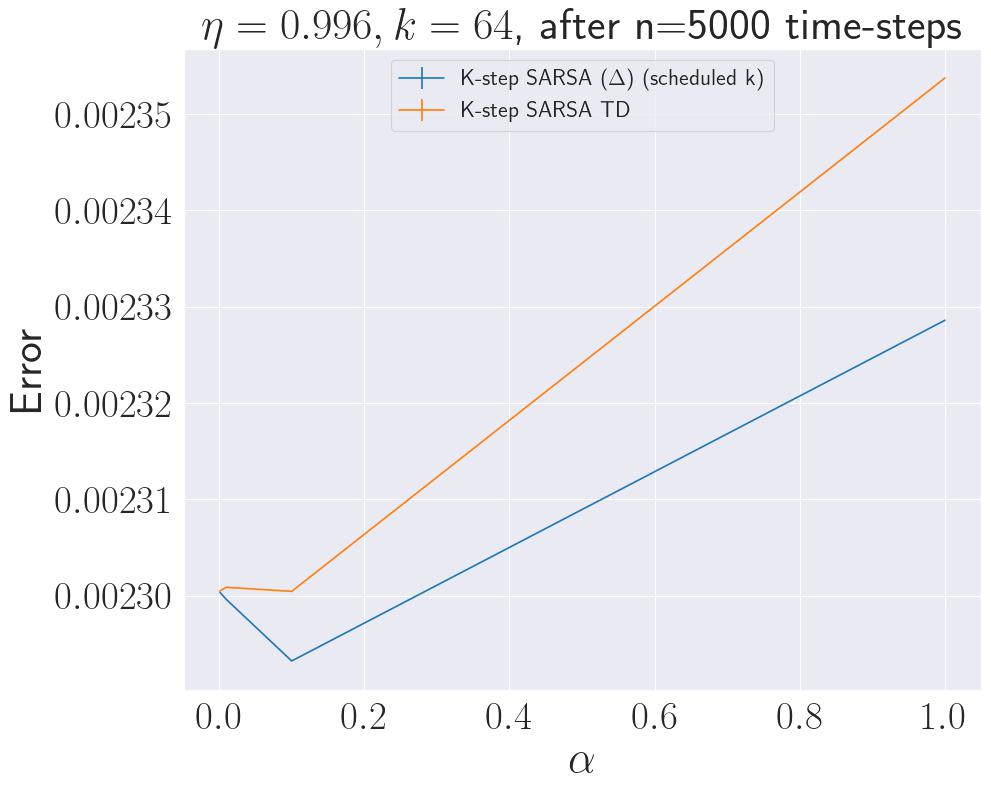}
\label{alpha}
}%
\subfloat[Error at different k values.]{
\includegraphics[width=0.5\columnwidth]{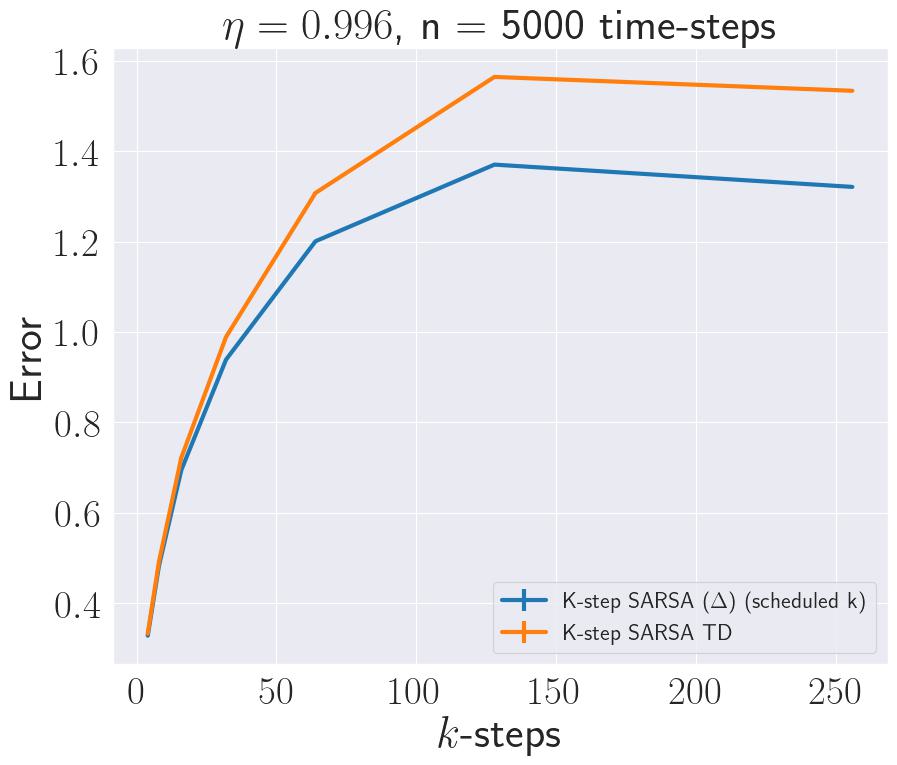}
\label{ksteps}
}%
\caption{Fig. \ref{alpha} illustrates the outcomes for $\eta_{M} = 0.996$ on the five-state Ring MDP. The error signifies the absolute discrepancy between the estimated and actual discounted Q-value function (precomputed via Q-value iteration), averaged across the complete learning trajectory of 5000 timesteps. Fig. \ref{ksteps} illustrates the average absolute error for the optimal learning rate at each k-step return, extending beyond the effective planning horizon of $\eta_{M}$.}
\label{alpha-step}
\end{figure}

The proposed method is contrasted with k-step SARSA, as described in Sutton et al.\cite[Chapter 7.2]{sutton1998introduction}. SARSA ($\Delta$) is evaluated against different RL benchmarks, including the Ring MDP \cite{kearns2000bias} and both deterministic and stochastic environments, to determine its efficacy. Fig. \ref{rmdp} illustrates the Ring MDP, which consists of a Markov process featuring five states arranged in a circular form. The probability of transitioning one state clockwise around the ring is 0.95, while the likelihood of remaining in the current state at each step is 0.05. In contrast to the other states on the ring that provide a reward of 0, two neighboring states present rewards of +1 and -1, respectively.
All experiments utilize the identical parameters established by Romoff et al.\cite{romoff2019separating}, which involve supplying a variable number of gamma values, commencing at 0 and incrementing according to $\eta_{m+1}=\frac{\eta_m + 1}{2}$ until the requisite maximum $\eta_{m}$ is attained. As previously mentioned in Section \ref{arv}, $K_{m} = \frac{1}{(1-\eta_{m})}$ for every m. The baseline consists of a singular estimator with $\eta = \eta_{m}$ and $k=k_{m}$. Fig. \ref{alpha} illustrates the effectiveness of two methodologies: K-step SARSA ($\Delta$) and K-step SARSA TD, utilizing a discount factor ($\eta$ = 0.996), k = 64 (a parameter related to the algorithm's step count), and various values of the learning rate $\alpha$. The y-axis represents the absolute error, while the x-axis indicates the parameter $\alpha$, which ranges from 0.0 to 1.0. At $\alpha$=0.0, both methods start with about the same amount of error (around 0.00230). As $\alpha$ increases, K-step SARSA ($\Delta$) exhibits a slight initial reduction in inaccuracy, followed by a consistent rise thereafter. In contrast, K-step SARSA TD shows a constant rise in error with low variability. Overall, the K-step SARSA ($\Delta$) approach outperforms K-step SARSA TD in terms of stability and accuracy across all $\alpha$ values. Both techniques are sensitive to fluctuations in $\alpha$, but K-step SARSA TD has a more significant rise in error.

Fig. \ref{ksteps} compares the efficacy of two approaches, K-step SARSA ($\Delta$) and K-step SARSA TD, based on the number of k-steps. The y-axis represents the absolute error, whereas the x-axis depicts k-steps, which relates to the number of steps utilized in the algorithms. The range of k-steps spans from 0 to approximately 250. K-step SARSA ($\Delta$) begins with a lower error at smaller k-steps, which then escalates with larger k-steps. K-step SARSA TD starts with an error level similar to that of K-step SARSA ($\Delta$) for lower k-steps; however, its error increases more sharply as k-steps increase. The overall pattern indicates that both techniques exhibit a rise in inaccuracy as the number of k-steps increases. K-step SARSA ($\Delta$) consistently demonstrates superior performance compared to K-step SARSA TD across all k-step values, evidenced by a lower error rate. The rapid error amplification noted in K-step SARSA TD indicates that the scheduling method employed in K-step SARSA ($\Delta$) successfully mitigates error accumulation. Additionally, in this particular MDP, the error increases with k as a result of the bias-variance trade-off linked to k-step returns, consistent with the findings of  Romoff et al.\cite{romoff2019separating}, Kearns and Singh\cite{kearns2000bias}.

Analysis of Figs. \ref{alpha} and \ref{ksteps}: In comparing the two figures, K-step SARSA ($\Delta$) consistently outperforms K-step SARSA TD in both scenarios ($\alpha$ and k-steps). Both approaches demonstrate sensitivity to increased parameter values ($\alpha$ or k-steps). K-step SARSA ($\Delta$) demonstrates improved stability and achieves lower error rates due to its scheduling methodology. Interestingly, compared to the growth in $\alpha$, errors climb faster with K-step SARSA TD as the number of k-steps increases.

\subsection{OpenAI Gym Environments}
\label{SecOpenAI}

Two algorithms, k-step SARSA($\Delta$) and k-step SARSA TD, were tested in the FrozenLake-v1, Taxi-v3, CliffWalking-v0, and Blackjack environments of OpenAI Gym. The results are shown in the figures below. The y-axis displays how many awards the algorithms got on average across episodes, while the x-axis shows how many episodes it took to train the algorithms. Episodes range from 0 to 100,000 episodes. The shaded area around each curve displays the standard error for 10 random seeds. Both algorithms utilize k = 32, which is the number of steps.

\begin{figure}[!htbp]
\centering
\subfloat[FrozenLake-v1.]{
\includegraphics[width=0.5\columnwidth]{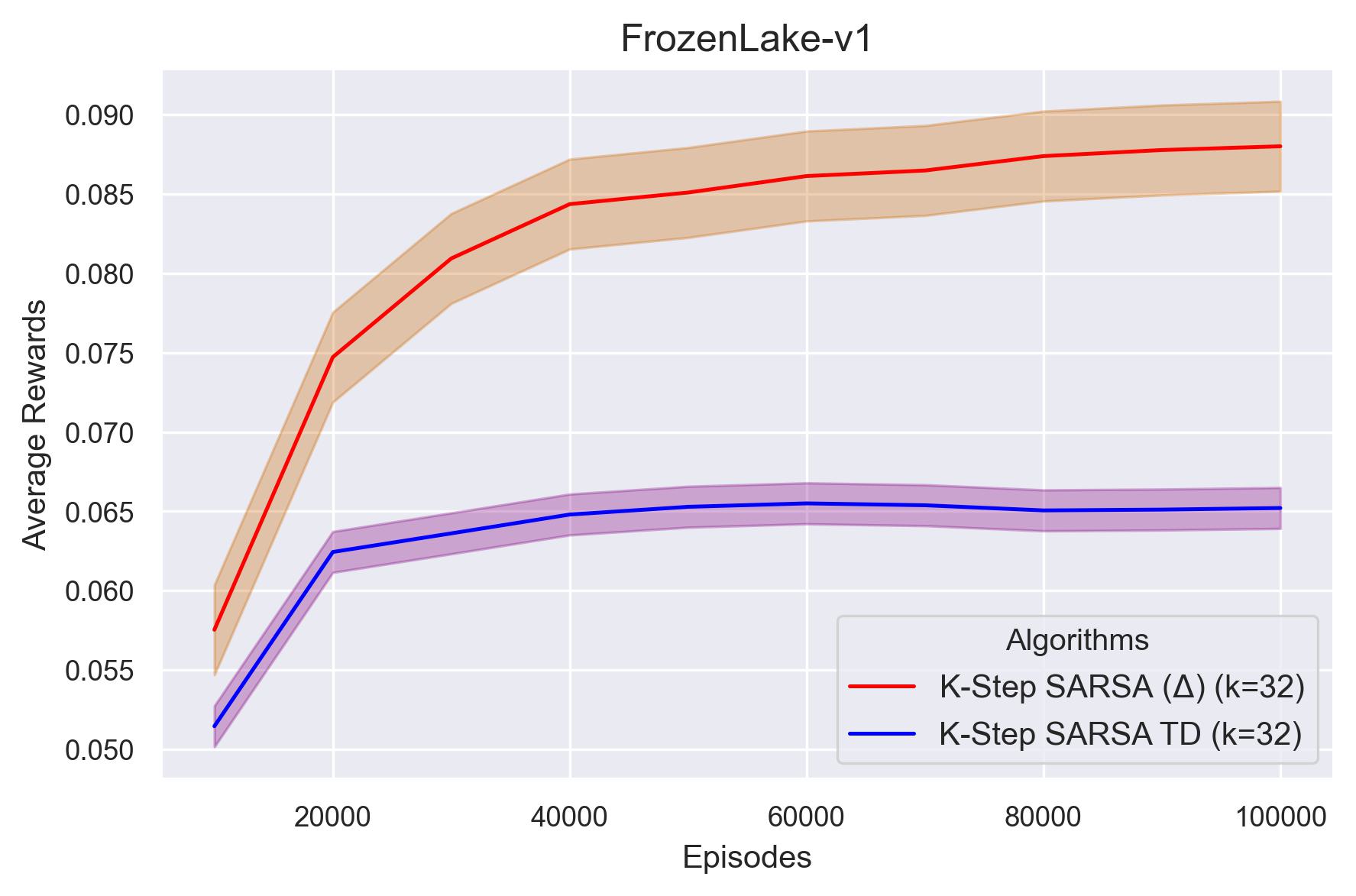}
\label{FrozenLake}
}%
\subfloat[Taxi-v3.]{
\includegraphics[width=0.5\columnwidth]{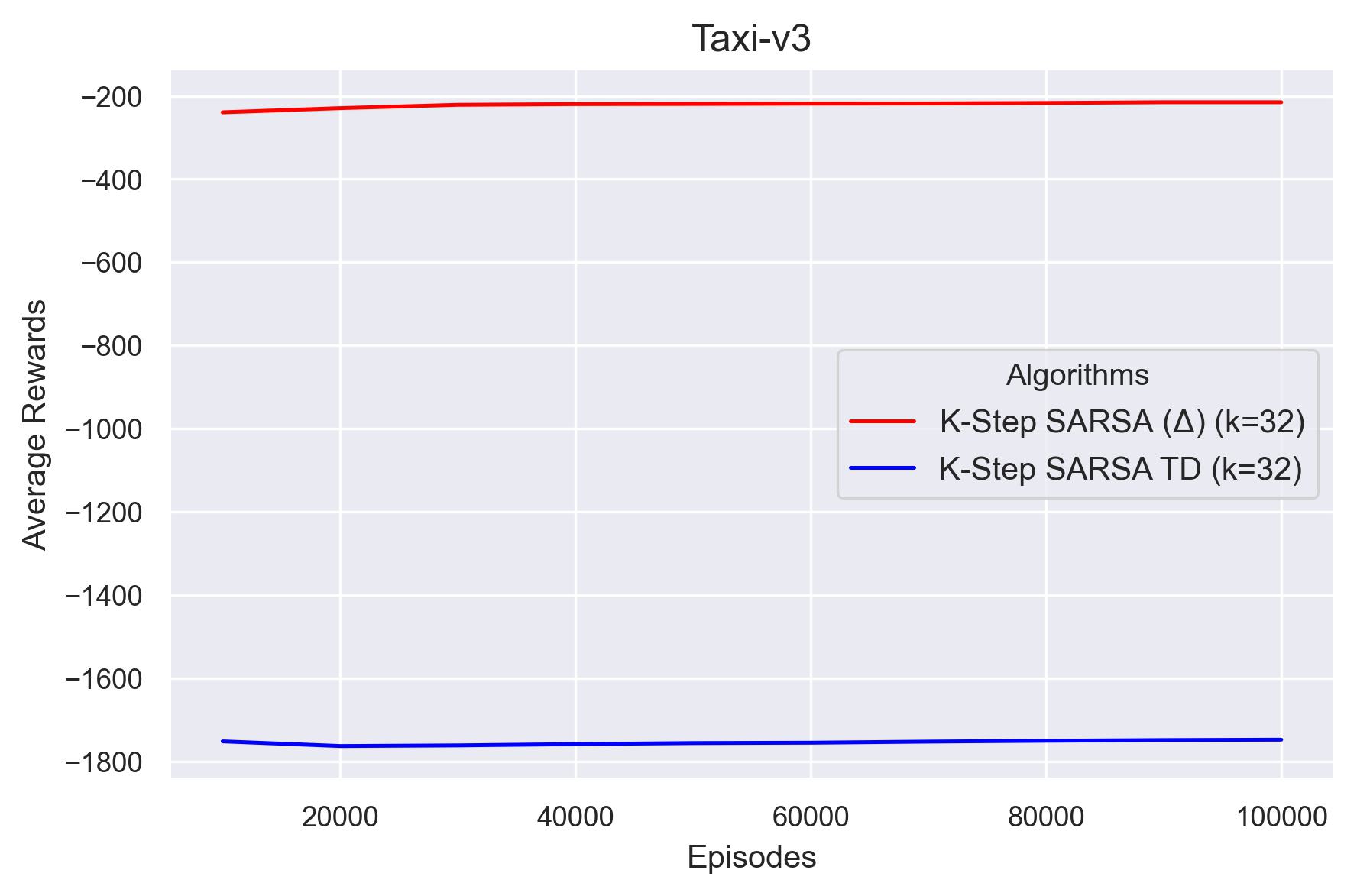}
\label{Taxi}
}%
\caption{Comparative analysis of K-step SARSA ($\Delta$) versus traditional K-step SARSA TD within the OpenAI Gym environments: FrozenLake-v1 and Taxi-v3. The shaded area illustrates the standard error calculated from 10 random seeds.}
\label{FrozenLake-Taxi}
\end{figure}

The red line denotes K-step SARSA ($\Delta$), while the blue line represents K-step SARSA TD in both Fig. \ref{FrozenLake} and Fig. \ref{Taxi}. Fig. \ref{FrozenLake} presents the FrozenLake-v1 environment, while Fig. \ref{Taxi} showcases the Taxi-v3 environment. Fig. \ref{FrozenLake} illustrates that k-step SARSA ($\Delta$) consistently achieves higher average rewards compared to k-SARSA TD as training progresses. Over the first 20,000 episodes, K-step SARSA ($\Delta$) shows fast learning with a significant rise in average rewards. However, k-step SARSA TD displays a smaller increase in average rewards when compared to k-step SARSA ($\Delta$) during the same episodes. After 40,000 episodes, k-step SARSA ($\Delta$) performance levels off and regularly beats k-step SARSA TD, showing that it is reliable over the long-term. k-step SARSA consistently performs worse than k-step SARSA ($\Delta$).

Fig. \ref{Taxi} illustrates that K-Step SARSA ($\Delta$) regularly achieves superior rewards (i.e., nearer to -200) during the training phase, signifying enhanced learning efficiency and overall performance. Fig. \ref{Taxi} also demonstrates that K-Step SARSA ($\Delta$) stabilizes in the initial phases, indicating robustness and the capacity to converge to an effective policy, whereas K-Step SARSA TD commences at considerably lower rewards (approximately -1800) and exhibits negligible improvement, remaining predominantly flat throughout the training process. The K-Step SARSA TD algorithm encounters difficulties in formulating an effective policy in this context relative to K-Step SARSA ($\Delta$). The results of the experiment indicate that K-Step SARSA ($\Delta$) demonstrates a significant advantage over K-Step SARSA TD in the Taxi-v3 environment, attaining considerably higher average rewards and achieving faster convergence.

\begin{figure}[!htbp]
\centering
\subfloat[CliffWalking-v0.]{
\includegraphics[width=0.5\columnwidth]{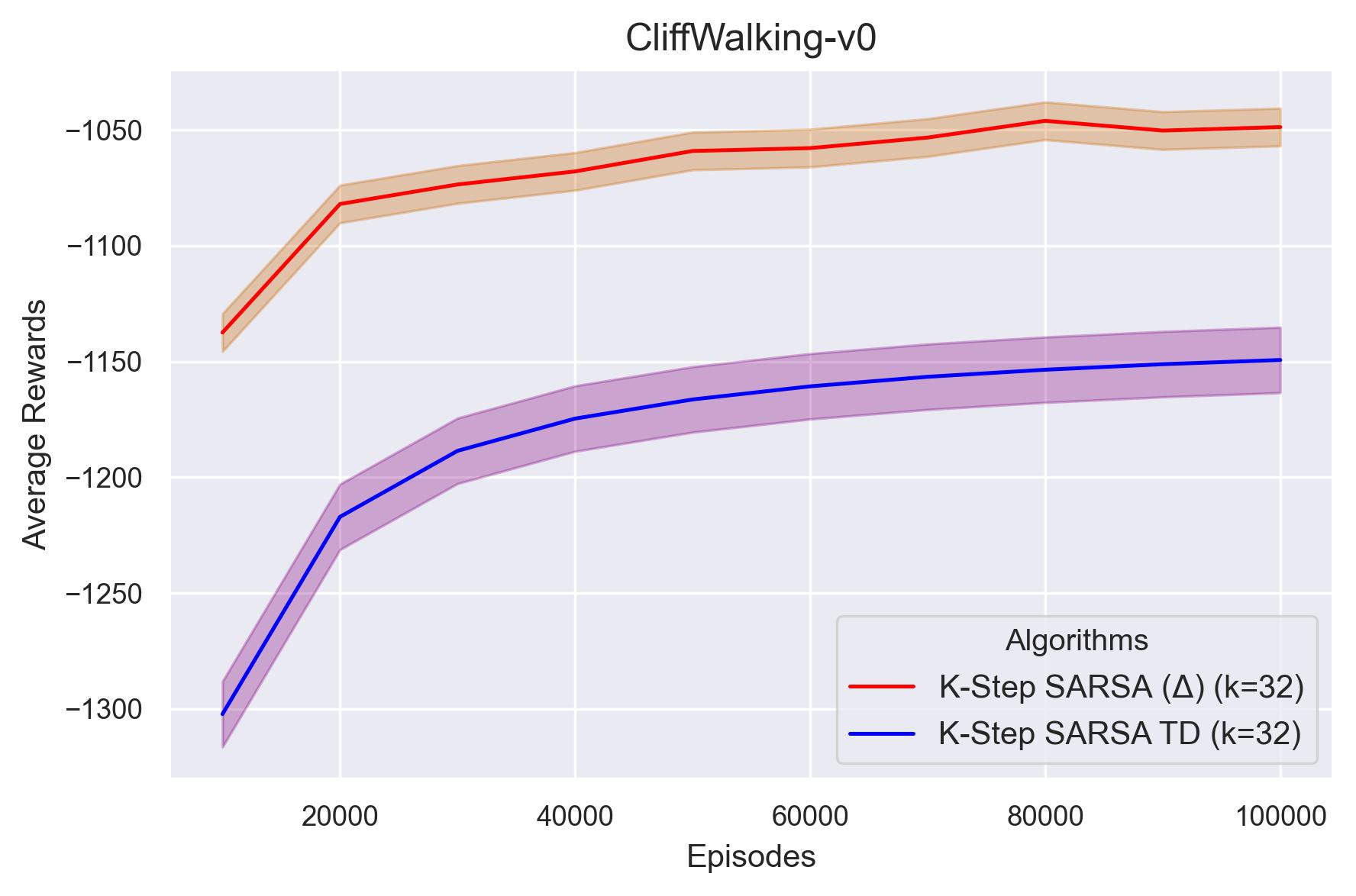}
\label{CliffWalking}
}%
\subfloat[Blackjack-v1.]{
\includegraphics[width=0.5\columnwidth]{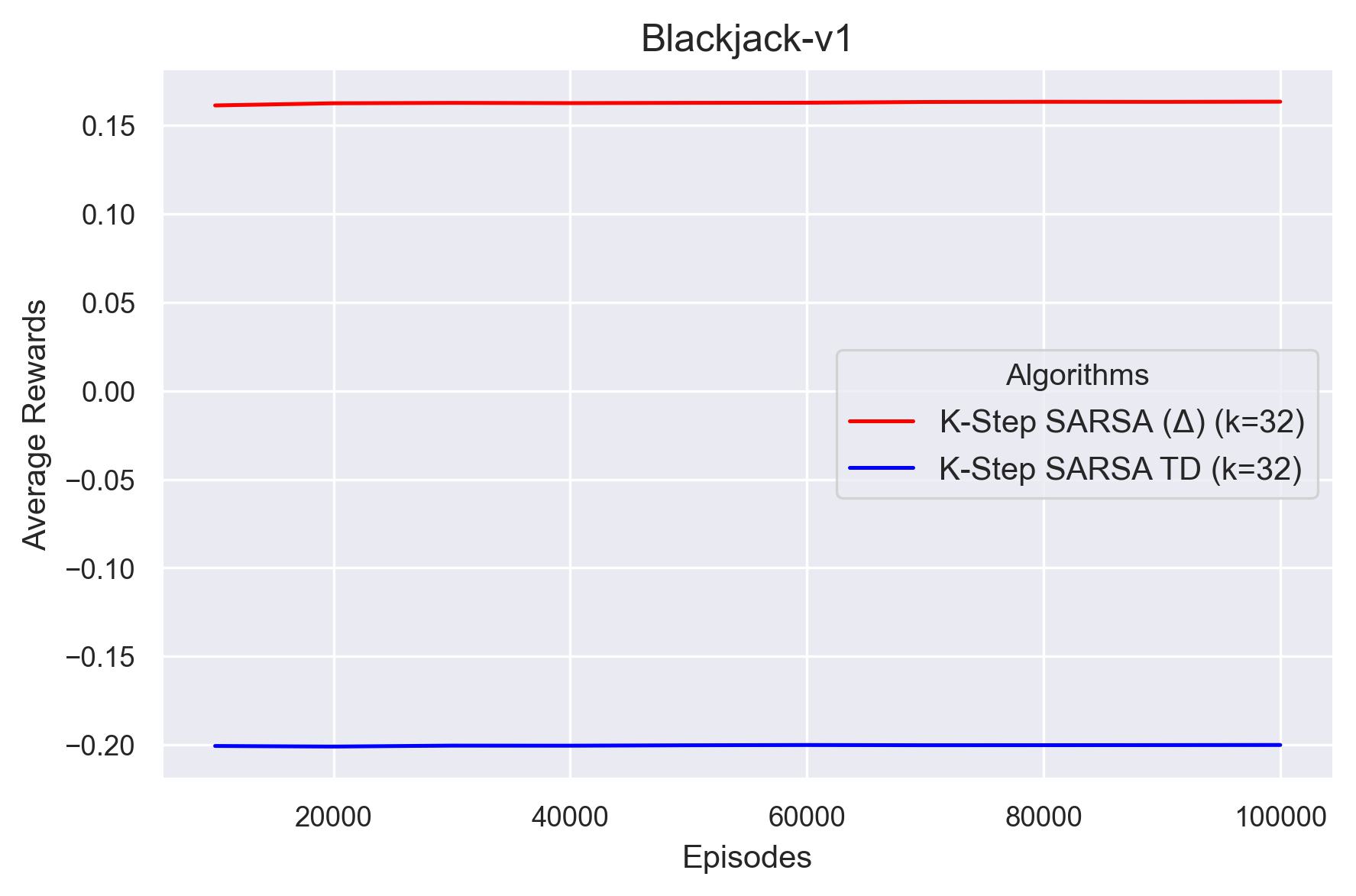}
\label{Blackjack}
}%
\caption{Comparative analysis of K-step SARSA ($\Delta$) versus traditional K-step SARSA TD within the OpenAI Gym environments: CliffWalking-v0 and Blackjack-v1. The shaded area illustrates the standard error calculated from 10 random seeds.}
\label{CliffWalking-Blackjack}
\end{figure}
Figs. \ref{CliffWalking} and \ref{Blackjack} demonstrate the learning effectiveness of two variations of the K-Step SARSA algorithm: K-Step SARSA ($\Delta$) and K-Step SARSA TD. This analysis is based on 100,000 episodes conducted in the CliffWalking-v0 and Blackjack-v1 environments. The evaluation of both algorithms is conducted with k=32, which is a hyperparameter indicating the number of steps involved in the updating process. The overall pattern of average rewards shown in Fig. \ref{CliffWalking} indicates that both algorithms exhibit an upward trajectory in average rewards as the number of episodes increases, reflecting enhanced policy performance through ongoing learning. Nonetheless, K-Step SARSA ($\Delta$) routinely exhibits better performance than K-Step SARSA TD, leading to increased average rewards across the episodes.

In comparison to K-Step SARSA TD, the red line in Fig. \ref{CliffWalking} representing K-Step SARSA ($\Delta$) shows a significant improvement in the first episodes (about 20,000 episodes), suggesting a faster convergence to optimum policies. K-Step SARSA ($\Delta$) receives a reward close to -1050 in the 100,000th episode, whereas K-Step SARSA TD receives a somewhat smaller reward of around -1120. This illustrates how K-Step SARSA ($\Delta$) performs better on this task.

The average reward of approximately +0.15 is dependably achieved across all episodes by the red curve in Fig. \ref{Blackjack} that represents K-Step SARSA ($\Delta$). The blue curve in Fig. \ref{Blackjack}, which represents K-Step SARSA TD, suggests stability; however, it presents an average reward that is significantly lower, at approximately -0.2, across all episodes. K-Step SARSA ($\Delta$) markedly surpasses K-Step SARSA TD regarding average rewards. The favorable average reward of K-Step SARSA ($\Delta$) demonstrates its superior suitability for policy optimization in the Blackjack-v1 environment, whereas the negative reward for K-Step SARSA TD signifies inferior performance.

In conclusion, the comparison of results across all four environments reveals a consistent advantage of K-Step SARSA ($\Delta$) over K-Step SARSA TD in each case. This advantage is demonstrated by increased average rewards, quicker convergence, and enhanced stability in learning results. Environments defined by deterministic state transitions, such as CliffWalking-v0 and FrozenLake-v1, clearly showcase the advantages of K-Step SARSA ($\Delta$). In contrast, more complex stochastic environments like Blackjack-v1 and Taxi-v3 further emphasize its robustness. The performance graph of deterministic environments, such as CliffWalking-v0 and FrozenLake-v1, resembles an exponential growth plateau curve due to predictable state transitions. On the contrary, the performance graph for sophisticated stochastic environments, such as Blackjack-v1 and Taxi-v3, resembles a flat line due to the uncertainty and unpredictability in state transitions.\par

Table~\ref{tab:k_step_sarsa_comparison} presents the average rewards along with confidence intervals (CI) for each algorithm within their respective environments. Kruskal statistical tests \cite{kruskal1952use} were conducted at a significance level of $\alpha=0.05$ to compare K-step SARSA TD and K-Step SARSA ($\Delta$) baseline models.
\begin{table}[!htbp]
\centering
\caption{This table presents a comparative analysis of K-Step SARSA TD and K-Step SARSA ($\Delta$) across multiple OpenAI Gym Toy Text environments.}
\resizebox{1.0\columnwidth}{!}{   
\begin{tabular}{|l|cccc|}
\hline
\textbf{Algorithm} & \multicolumn{4}{c|}{\textbf{Environments}} \\
\cline{2-5}
&\textbf{Blackjack-v1} & \multicolumn{1}{|c}{\textbf{CliffWalking-v0}} & \multicolumn{1}{|c}{\textbf{FrozenLake-v1}} & \multicolumn{1}{|c|}{\textbf{Taxi-v3}} \\
\hline
K-Step SARSA ($\Delta$) & 0.16$\pm$0.00011 & \multicolumn{1}{|c}{-1067.66$\pm$5.041} & \multicolumn{1}{|c}{0.081$\pm$0.0017} & \multicolumn{1}{|c|}{-189.96$\pm$1.39} \\
\hline
K-Step SARSA TD & 0.20$\pm$0.000052 & \multicolumn{1}{|c}{-1182.07$\pm$8.75} & \multicolumn{1}{|c}{0.063$\pm$0.00079} & \multicolumn{1}{|c|}{-1754.70$\pm$0.97} \\
\hline
\end{tabular}
}
\label{tab:k_step_sarsa_comparison}
\end{table}


\subsection{OpenAI Gym Atari Environments}
\label{secAtari}
\begin{figure}[!htbp]
\centering
\subfloat[SeaquestNoFrameskip-v4]{
\includegraphics[width=0.5\columnwidth, height=0.35\columnwidth]{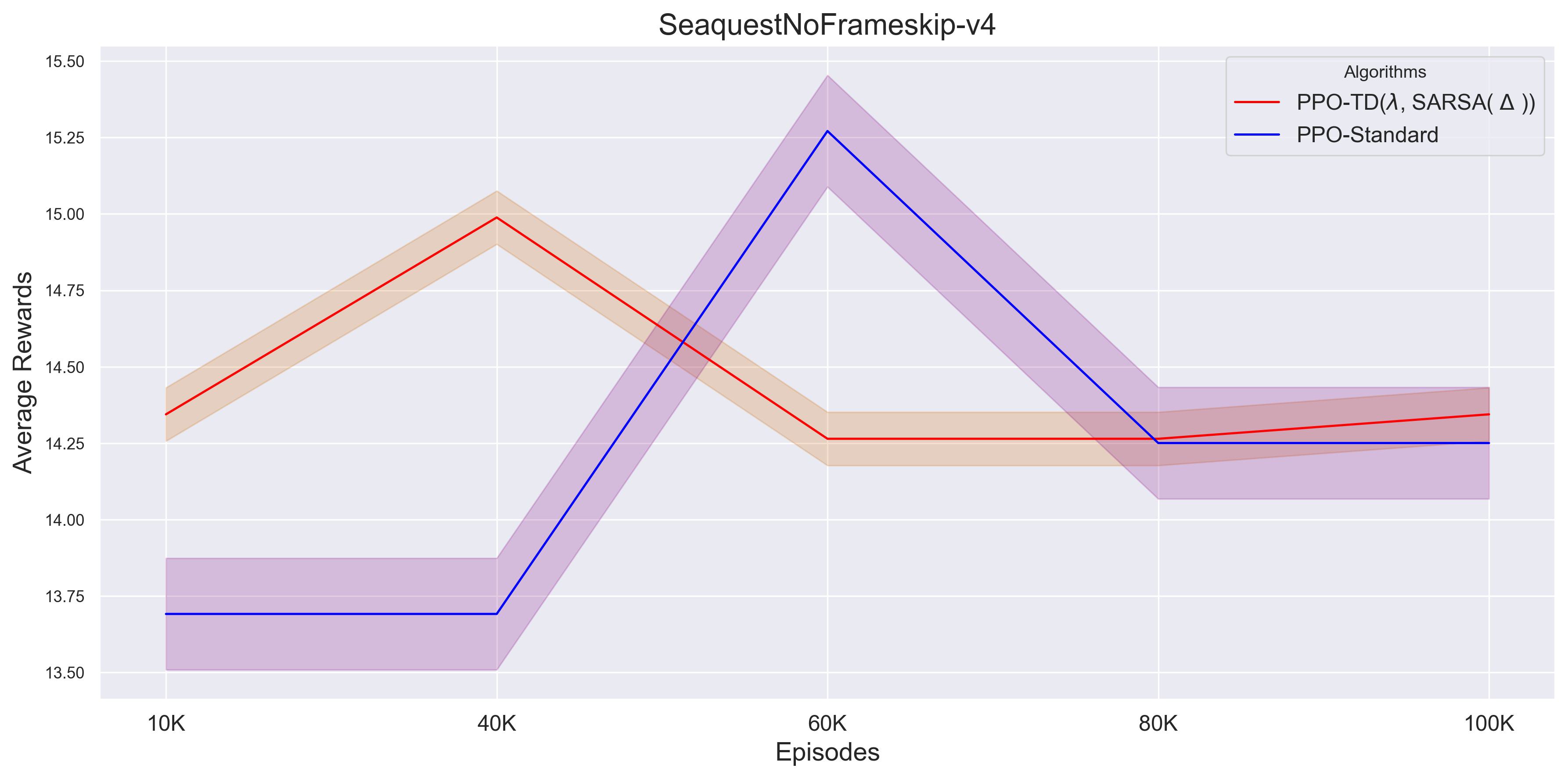}
\label{Seaquest-res}
}%
\subfloat[QbertNoFrameskip-v4]{
\includegraphics[width=0.5\columnwidth, height=0.35\columnwidth]{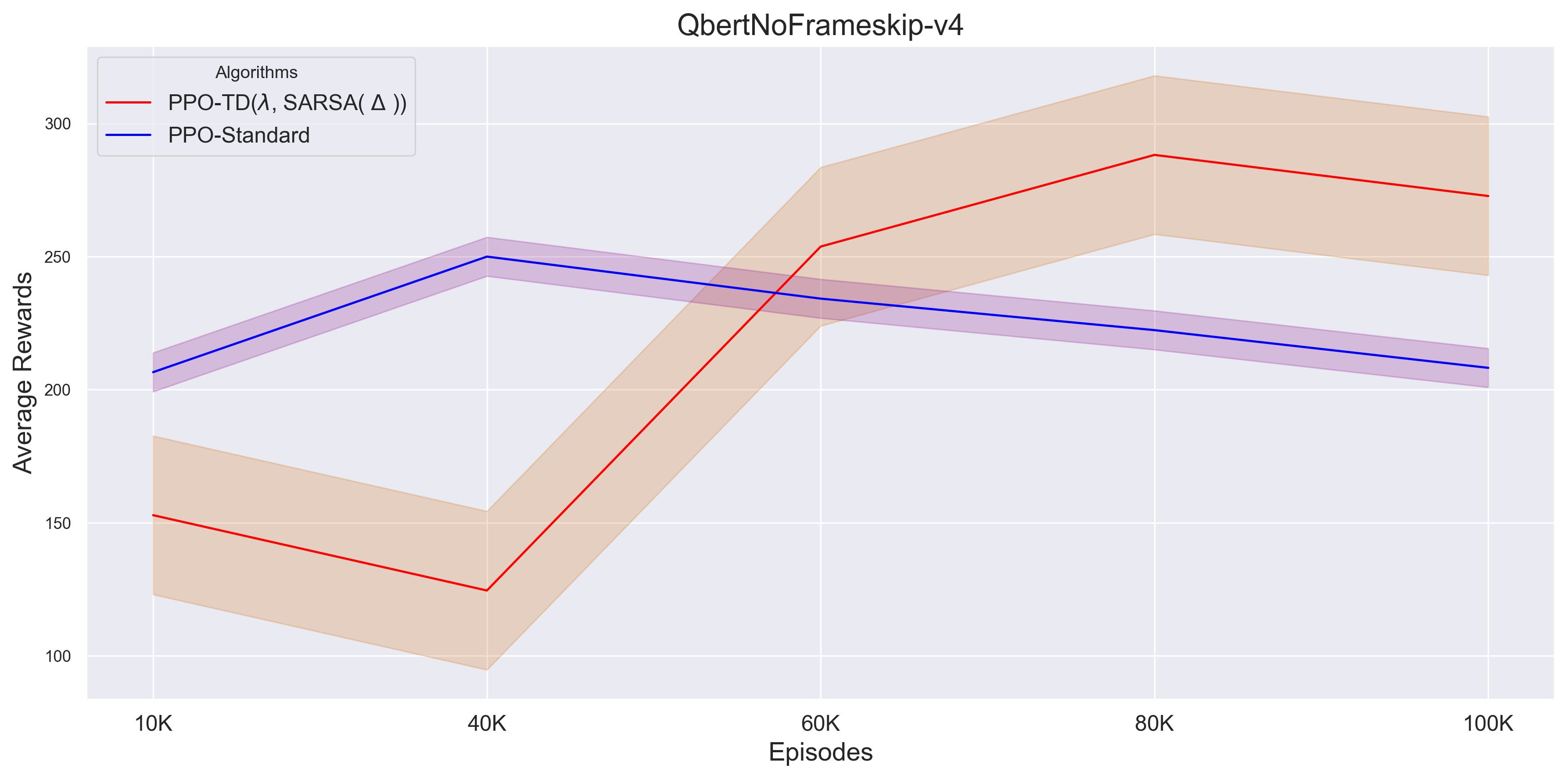}
\label{Qbert-res}
}%
\caption{
Comparison of performance between PPO-Standard and PPO-TD($\lambda$, SARSA($\Delta$)) across two Atari games: \textbf{(a)} SeaquestNoFrameskip-v4 and \textbf{(b)} QbertNoFrameskip-v4. The x-axis denotes the quantity of training episodes (measured in thousands), whereas the y-axis illustrates the average rewards obtained. The shaded areas represent the standard deviation calculated from 10 runs with different random seeds.}
\label{fig:qbert-seaquest}

\end{figure}

To assess the efficacy of the proposed PPO-TD($\lambda$, SARSA($\Delta$)) method, experiments were conducted in two Atari environments: \textit{SeaquestNoFrameskip-v4} and \textit{QbertNoFrameskip-v4} \cite{bellemare2016unifying}, and the results were compared with those obtained from the standard PPO algorithm \cite{schulman2017proximal}. The results shown in Figs.~\ref{Seaquest-res} and ~\ref{Qbert-res} show that in both environments, the proposed technique achieves better stability and better long-term rewards. The experiments tracked average rewards during training episodes ranging from 10,000 to 100,000. The experiments employ the code cited in Kostrikov\cite{pytorchrl}.\par 
In \textit{SeaquestNoFrameskip-v4}, PPO-TD($\lambda$, SARSA($\Delta$)) demonstrates superior performance compared to PPO-Standard at four out of five evaluation points, as illustrated in Fig.~\ref{Seaquest-res}. Following 100K episodes, the mean reward for PPO-TD($\lambda$, SARSA($\Delta$)) is around \textbf{14.35}, while PPO-Standard attains \textbf{14.25}, reflecting a \textbf{0.7\% enhancement}. PPO-Standard has a pronounced peak after 60K episodes, achieving an average reward of around 15.25, after which it declines and plateaus. In contrast, PPO-TD($\lambda$, SARSA($\Delta$)) exhibits more refined learning dynamics and sustains stable performance following its peak at 40K episodes.\par

The performance difference in \textit{QbertNoFrameskip-v4} is more substantial, as illustrated in Fig.~\ref{Qbert-res}. At 100K episodes, PPO-TD($\lambda$, SARSA($\Delta$)) attains an average reward of approximately \textbf{275}, whereas PPO-Standard achieves about \textbf{210}, leading to an improvement of approximately \textbf{31\%}. It is noteworthy that PPO-TD($\lambda$, SARSA($\Delta$)) starts with lower rewards, around \textbf{155} at 10K episodes, in contrast to \textbf{210} for PPO-Standard. However, it exhibits a steady progress, finally surpassing the baseline after 60K episodes, and reaching a high of around \textbf{290} episodes after 80K episodes. The wider shaded region for the proposed technique in the earlier phases indicates more variance, which diminishes as training progresses—signifying improved convergence and learning efficiency with time.\par

In conclusion, these results highlight the advantages of integrating PPO and SARSA($\Delta$) with TD($\lambda$, $\Delta$). The suggested method enhances the stability of action-value estimates across various time scales and fosters superior long-term performance in dense-reward environments, especially in intricate games such as \textit{Qbert}.

Table~\ref{tab:ppo_td_sarsa_comparison} shows the average rewards for each algorithm in their own environments, along with confidence intervals (CI). We used Kruskal statistical tests \cite{kruskal1952use} at a significant level ($\alpha=0.05$) to compare the PPO-TD($\lambda$, SARSA($\Delta$))/PPO baseline models.
\begin{table}[!htbp]
\centering
\caption{Comparative analysis of PPO and PPO-TD($\lambda$, SARSA($\Delta$)) across various OpenAI Gym Atari environments.}
\begin{tabular}{|l|cc|}
\hline
\textbf{Algorithm} & \multicolumn{2}{c|}{\textbf{Atari Environments}} \\
\cline{2-3}
& \textbf{Seaquest} & \multicolumn{1}{|c|}{\textbf{Qbert}} \\
\hline
PPO-TD ($\lambda$, SARSA ($\Delta$)) & 14.44$\pm$0.076 & \multicolumn{1}{|c|}{218.42$\pm$18.46} \\
\hline
PPO & 14.23$\pm$0.15 & \multicolumn{1}{|c|}{224.27$\pm$4.53} \\
\hline
\end{tabular}
\label{tab:ppo_td_sarsa_comparison}
\end{table}

\section{Discussion and Conclusions}
\label{conclusion}
By combining SARSA with TD($\Delta$) and various discount factors, SARSA($\Delta$) improves SARSA's stability and performance. This method maintains the benefits of on-policy learning while enabling a more precise breakdown of action-values. To evaluate the effectiveness of SARSA($\Delta$), experiments were performed in both deterministic and stochastic environments from OpenAI Gym, as well as in a synthetic tabular environment—specifically, the five-star ring MDP. Furthermore, enhancements in performance within dense-reward environments of the Atari suite were demonstrated through a PPO-based variant of SARSA($\Delta$). The experimental results across all examined environments indicate that the proposed method, SARSA($\Delta$), surpasses k-step SARSA TD by achieving higher average rewards, quicker convergence, and more consistent learning trajectories. The PPO-based variant of SARSA ($\Delta$) also demonstrates superior performance compared to standard PPO algorithms in dense-reward Atari environments, as evidenced by the experimental results outlined in Section \ref{secAtari}. This research contributes to our knowledge of multi-step RL algorithms and its applicability to diverse sequential decision-making tasks.\\
\textit{Limitations}: The proposed SARSA($\Delta$) method has four limitations that require attention:
(i) Memory Overhead: When using a large number of $\Delta$ components, the memory needs increase since the component-level updates scale linearly with the number of discount factors. This technique requires an extra O(M) in storage compared to conventional SARSA. (ii) Stochastic Environments: Best results may not be attained through rigid scheduling of $\eta_{m}$ in scenarios characterized by high unpredictability of rewards. In certain cases, adaptive scheduling of $\eta_{m}$ can boost both the stability and the effectiveness of learning. (iii) Partial Observability Markov Decision Process (POMDP): Current formulations are inapplicable to environments characterized by partial observability, as they presuppose full observability of the Markov Decision Process (MDP). (iv) Extension to off-policy: There are numerous issues associated with adapting SARSA($\Delta$) to off-policy algorithms such as Q-learning. Off-policy learning has the potential to significantly increase variance in long-horizon components, which may subsequently lead to instability during the training process.\\
\textit{Future Directions}: (i) Incorporation with Off-Policy Q-Learning: The adaptability of the SARSA($\Delta$) idea might be enhanced by applying it to off-policy algorithms, similar to the Q-learning version of TD($\Delta$). (ii) Applications to Multi-Agent Systems: In a multi-agent system, the agents coordinate on aligned subgoals by sharing $D_{m}$ components. (iii) Adaptive Scheduling via Meta-Gradients: Performance in various environments may be improved by using meta-gradient approaches for dynamically adjusting $\eta_{m}$. (iv) Extension to Partially Observable: Adapting SARSA($\Delta$) for environments characterized by partial observability is an effective approach to enhance its relevance in real-world applications. We decided to take care of it later.

\bibliography{main.bib}

\section*{Acknowledgments}
This research is funded by the Innovation Teams of Ordinary Universities in Guangdong Province under Grants (2021KCXTD038, 2023KCXTD022), the Key Laboratory of Ordinary Universities in Guangdong Province (2022KSYS003), Key Discipline Research Ability Improvement Project of Guangdong Province (2021ZDJS043, 2022ZDJS068), the Natural Science Foundation of Guangdong Province (2022A1515010990), the Research Fund of the Department of Education of Guangdong Province Grants (2022KTSCX079, 2023ZDZX1013, 2022ZDZX3012, 2022ZDZX3011, 2023ZDZX2038), Chaozhou Engineering Technology Research Center (z25025) and Hanshan Normal University project (XY202105). We express our gratitude to the editors and referees for their invaluable suggestions and remarks.

\section*{Author contributions statement}
M.H. participated in all stages, from Conceptualization design to writing---original draft preparation. G.Z., and X.D. funding acquisition. X.C. and Y.Z. supervision and project administration. W.H., L.M., S.Q., Z.Z., and P.W. investigation. Z.U. and N.J. validation and formal analysis.

\section*{Data availability}
The corresponding authors may be contacted to request data related to this paper.
\section*{Competing interests}
The authors disclose no conflicting interests.
\pagebreak

\section*{Appendix}
\appendix

\section[\appendixname~\thesection]{PROOFS}
\label{proofs}
\begin{proof}{Theorem \ref{thm:thm1}}
\begingroup
\allowdisplaybreaks
\begin{gather}
\begin{split}
\shortintertext{The proof utilizes induction. Initially, it is necessary to establish that the base case is valid at n=0. This statement is evidently valid given our initial assumption, particularly in the context of zero initialization. At a specific time-step n, we assume the validity of the statement, namely, $\sum_{m=0}^{M}\theta_{n}^{m} = \theta_{n}^{\eta}$. We will now demonstrate that it also holds at the subsequent time step, n+1.}\nonumber
\sum_{m=0}^{M}\theta_{n+1}^{m} &= \sum_{m=0}^{M}\bigg( \theta_{n}^{m} + \alpha_{m} \bigg(G_{n}^{M, \lambda_{m}} - \hat{D}_{m}(s_{n}, a_{n})\bigg)\phi(s_{n},a_{n})\bigg)\\ \nonumber
   D_{m}(s, a) &= Q_{\eta_{m}}(s, a) \mathit{-} Q_{\eta_{m -1}}(s, a) \text{ and using Eq. \ref{SRD-Estimator}}\\ \nonumber
               &= \theta_{n}^{\eta} + \sum_{m=0}^{M}\alpha_{m} \bigg(\sum_{k=n}^{\infty}(\lambda_{m}\eta_{m})^{k-n}\delta_{k}^{m}\bigg)\phi(s_{n},a_{n})\text { , Assumption based on induction.}\\ \nonumber
               &= \theta_{n}^{\eta} + \alpha \sum_{k=n}^{\infty}(\lambda \eta)^{k-n} \underbrace{\bigg( \sum_{m=0}^{M}\delta_{k}^{m}\bigg)}_{*}\phi(s_{n},a_{n})\text{ , Because of $\alpha_{m} = \alpha$, $\lambda_{m}\eta_{m} = \lambda_{\eta}$, $\forall m$}
\shortintertext{To show that $\sum_{m=0}^{M}\theta_{n+1}^{m} = \theta_{n+1}^{\eta}$, we must establish that the term (*) = $\sum_{m=0}^{M}\delta_{k}^{m}$ is equivalent to the conventional TD error $\delta_{k}^{\eta}$.}\nonumber
\sum_{m=0}^{M}\delta_{k}^{m} &= r_{k} +\eta_{0}\hat{Q}_{\eta_{0}}(s_{k+1},a_{k+1}) - \hat{Q}_{\eta_{0}}(s_{k},a_{k}) + \sum_{m=1}^{M}\bigg( (\eta_{m} - \eta_{m-1})\sum_{q=0}^{m-1}\langle\theta_{n}^{q},\phi(s_{k+1},a_{k+1})\rangle \\\nonumber&+ \eta_{m}\langle\theta_{n}^{m},\phi(s_{k+1},a_{k+1})\rangle - \langle\theta_{n}^{m},\phi(s_{k},a_{k})\rangle \bigg)\\ \nonumber
&= r_{k} +\eta_{0}\hat{Q}_{\eta_{0}}(s_{k+1},a_{k+1}) + \bigg\langle \sum_{m=1}^{M}\bigg(\eta_{m} \sum_{q=0}^{m}\theta_{n}^{q} - \eta_{m-1}\sum_{q=0}^{m-1}\theta_{n}^{q}\bigg),\phi(s_{k+1},a_{k+1})\bigg\rangle \\\nonumber&- \bigg\langle \sum_{m=0}^{M}\theta_{n}^{m},\phi(s_{k},a_{k})  \bigg\rangle \\ \nonumber
&= r_{k} +\eta_{0}\hat{Q}_{\eta_{0}}(s_{k+1},a_{k+1})+ \eta_{m} \bigg\langle \sum_{m=0}^{M}\theta_{n}^{m},\phi(s_{k+1},a_{k+1})\bigg\rangle - \eta_{0}\bigg\langle\theta_{n}^{0},\phi(s_{k+1},a_{k+1})\bigg\rangle \\\nonumber&- Q_{\eta}(s_{k},a_{k})\\ \nonumber
&= r_{k} +\eta_{0}\hat{Q}_{\eta_{0}}(s_{k+1},a_{k+1}) + \eta \hat{Q}_{\eta}(s_{k+1},a_{k+1}) - \eta_{0}\hat{Q}_{\eta_{0}}(s_{k+1},a_{k+1}) - \hat{Q}_{\eta}(s_{k},a_{k})\\ \nonumber
&= r_{k} + \eta \hat{Q}_{\eta}(s_{k+1},a_{k+1}) - \hat{Q}_{\eta}(s_{k},a_{k})\\
&= \delta_{k}^{\eta}
\end{split}
\end{gather}
\endgroup
\end{proof}

\begin{proof}{Theorem \ref{thm:thm2}}
\begingroup
\allowdisplaybreaks
\begin{align}
\shortintertext{The following is the definition of the Bellman operator:} \nonumber
\mathrm{T} = r + \eta P \\ \nonumber
\shortintertext{The discount factor, the expected reward function, and the transition probability operator that is induced by the policy $\pi$ are denoted by the symbols $\eta$, r, and P, respectively. A geometric weighted sum of $\mathrm{T}$ is the definition of the TD($\lambda$) operator, which can be expressed as follows:}
\mathrm{T} = (1-\lambda)\sum_{k=0}^{\infty} \lambda^{k}(\mathrm{T})^{k+1}\label{SARSA-Bellman-Operator}\\
\shortintertext{It may be inferred that $\lambda \in [0, 1]$ is necessary for the aforementioned sum to be finite. A corresponding definition of $\mathrm{T}_{\lambda}$ is as follows for every function D:}
\mathrm{T}_{\lambda}D = D + (I - \lambda \eta P)^{-1}(\mathrm{T}D - D)\label{SARSA-Sep-Action-Value}\\
\shortintertext{The separated action-value function is denoted by D in this form, and the update is given in terms of the operator $\mathrm{T}_{\lambda}$, which is responsible for applying time-scale separation. In the event that $0\leq \lambda\eta<1$, the formula is considered to be well-defined. This is because the spectral norm of the operator $\lambda\eta P$ is less than 1, and thus, Eq. $I - \lambda\eta P$ is invertible. On the other hand, when the value of $\lambda$ is bigger than one, the equivalence between Eqs. \ref{SARSA-Bellman-Operator} and \ref{SARSA-Sep-Action-Value} is lost. This is not a problem because, in fact, training is based on the TD error, which, according to expectations, corresponds to the definition of $\mathrm{T}_{\lambda}$ that is provided in Eq. \ref{SARSA-Sep-Action-Value}. Now, let's have a look at the contraction property that the operator $\mathrm{T}_{\lambda}$ possesses. To begin, the Eq. designated as \ref{SARSA-Sep-Action-Value} can be rewritten as follows:}
\mathrm{T}_{\lambda} = (I - \lambda \eta P)^{-1}(\mathrm{T}D - \lambda \eta P D)\\ \nonumber
\shortintertext{The contraction property is demonstrated by examining two action-value functions, $D_{1}$ and $D_{2}$. The distinction between the updates of $\mathrm{T}_{\lambda}D_1$ and $\mathrm{T}_{\lambda}D_2$ is:} \nonumber
\mathrm{T}_{\lambda}D_{1} - \mathrm{T}_{\lambda}D_{2} = (I - \lambda \eta P)^{-1}\bigg(\mathrm{T}D_{1} - \mathrm{T}D_{2} - \lambda \eta P (D_{1} - D_{2})\bigg)\\ \nonumber
= (I - \lambda \eta P)^{-1}\bigg(\eta P (D_{1} - D_{2}) - \lambda \eta P (D_{1} - D_{2})\bigg)\\ \nonumber
= (I - \lambda \eta P)^{-1}\bigg(\eta (1-\lambda)P (D_{1} - D_{2})\bigg)\\ \nonumber
\shortintertext{The following is derived:} \nonumber
\|\mathrm{T}_{\lambda}D_{1} - \mathrm{T}_{\lambda}D_{2}\| \leq \frac{\eta|1 - \lambda|}{1 - \lambda\eta}\|D_{1} - D_{2}\| \textit{ where P = 1}\\ \nonumber
\shortintertext{It is established that $0\leq \lambda\leq 0$ constitutes a contraction. For $\lambda > 1$, the following condition must be satisfied:} \nonumber
\frac{\eta(\lambda - 1)}{1 - \lambda\eta} < 1 \Rightarrow \eta\lambda - \eta < 1 - \lambda\eta\\
\Rightarrow 2 \lambda\eta < 1 + \eta \Rightarrow \lambda < \frac{1 + \eta}{2\eta}\\
\shortintertext{Consequently, $\mathrm{T}_{\lambda}$ constitutes a contraction if $0 \leq \lambda < \frac{1 + \eta}{2\eta}$. This directly indicates that for $\eta < 1$, $\eta\lambda < 1$.} \nonumber
\end{align}
\endgroup
\end{proof}

\begin{proof}{Theorem \ref{thm:thm3}}
\begingroup
\allowdisplaybreaks
\begin{align}\label{prf:thm3}\nonumber
\shortintertext{In SARSA, the $\mathit{Q-value}$ function, Q(s, a), is estimated using samples of state-action pairs and their corresponding rewards. Let $Q_{n}(s,a)$ denote the estimate of the $\mathit{Q-value}$ at time n. For $\mathit{n}$ samples, by Hoeffding's inequality provides a guarantee for a variable that is bounded between $[-1, +1]$, that, the probability that the sample mean deviates from the true mean by more than $\epsilon$ is expressed as follows:}\nonumber
Q^{\pi}(s, a) = \mathbf{E}\bigg[\eta_{0} + \eta r_{1} + ... + \eta^{k-1}r_{k-1} + \eta^{k}Q^{\pi}(s_{k}, a_{k}) \bigg]\nonumber
\shortintertext{Here, the expectations are over a random trajectory under $\pi$; thus $\mathbf{E}[r_{\iota}](\iota \leq k \mathit{-}1)$ denotes the expected value of the $\iota$th reward received, while $\mathbf{E}[Q^{\pi}(s_{k}, a_{k}) ]$ is the expected value of the true value function at the $k$th state-action reached. The phased TD(k) update sums the terms $\eta^{\iota}\bigg( \frac{1}{n}\bigg)\sum_{i=1}^{n}r_{\iota}^{i}$, whose expectations are exactly the $\eta^{\iota}\mathbf{E}[r_{\iota}]$ appearing above \cite{kearns2000bias}.}\nonumber
\mathbb{P}\bigg(\bigg|\frac{1}{n}\sum_{x=1}^{n} r_{i}^{(x) } \mathit{-} \mathbb{E}[r_{i}] \bigg|\geq\epsilon\bigg) \leq 2e^{\bigg(\mathit{-} \frac{2n^{^{2}}\epsilon^{2}}{\sum_{x=1}^{n}(b \mathit{-} a)^{2}} \bigg)}\textit{where a=-1 and b=+1}\nonumber\\
\mathbb{P}\bigg(\bigg|\frac{1}{n}\sum_{x=1}^{n} r_{i}^{(x) } \mathit{-} \mathbb{E}[r_{i}] \bigg|\geq\epsilon\bigg) \leq 2e^{\bigg(\mathit{-} \frac{2n^{^{2}}\epsilon^{2}}{n2^{2}} \bigg)}\\
\shortintertext{Assuming that n and the probability of surpassing an $\epsilon$ value is fixed to be atmost $\delta$, we can derive the corresponding value of $\epsilon$:}\nonumber
2e^{\bigg(\mathit{-} \frac{2n^{^{2}}\epsilon^{2}}{n2^{2}} \bigg)} = \delta \nonumber\\
e^{\bigg(\mathit{-} \frac{2n^{^{2}}\epsilon^{2}}{n2^{2}} \bigg)} = \frac{\delta}{2}\nonumber\\
\shortintertext{Taking the natural logarithm}\nonumber
ln\bigg(e^{\bigg(\mathit{-} \frac{2n^{^{2}}\epsilon^{2}}{n2^{2}} \bigg)}\bigg) = ln\bigg(\frac{\delta}{2}\bigg)\nonumber\\
\mathit{-} \frac{2n^{^{2}}\epsilon^{2}}{n2^{2}} = ln\frac{\delta}{2}\nonumber\\
\mathit{-} \frac{n\epsilon^{2}}{2} = ln\frac{\delta}{2}\nonumber\\
\mathit{-} \frac{n\epsilon^{2}}{2} = ln\frac{\delta}{2}\nonumber\\
\frac{n\epsilon^{2}}{2} = \mathit{-}ln\frac{\delta}{2}\nonumber\\
\frac{n\epsilon^{2}}{2} = \mathit{-}\bigg(ln(\delta) - ln(2)\bigg)\nonumber\\
\frac{n\epsilon^{2}}{2} = \mathit{-}\bigg(ln(\delta) + ln(2)\bigg)\nonumber\\
\frac{n\epsilon^{2}}{2} = \bigg(ln(2) \mathit{-} ln(\delta)\bigg)\nonumber\\
\frac{n\epsilon^{2}}{2} = ln\frac{2}{\delta}\nonumber\\
n\epsilon^{2} = 2 log\frac{2}{\delta}\nonumber\\
\epsilon^{2} = \frac{2 log\frac{2}{\delta}}{n}\nonumber\\
\epsilon = \sqrt{\frac{2 log\frac{2}{\delta}}{n}}\\
\shortintertext{So by Hoeffding's inequality, for n samples, the following holds with probability at least $1 \mathit{-} \delta$,}\nonumber\\
\mathbb{P}\bigg(\bigg|\frac{1}{n}\sum_{x=1}^{n} r_{i}^{(x) } \mathit{-} \mathbb{E}[r_{i}] \bigg|\bigg) \leq \epsilon = \sqrt{\frac{2 log\frac{2}{\delta}}{n}}\\
\shortintertext{Since the analysis involves k different state-action pairs, a union bound is applied. To ensure the probability holds for all k hypotheses, adjust $\delta$ to $\delta/k$:}\nonumber
\mathbb{P}\bigg(\bigg|\frac{1}{n}\sum_{x=1}^{n} r_{i}^{(x) } \mathit{-} \mathbb{E}[r_{i}] \bigg|\bigg) \leq \epsilon = \sqrt{\frac{2 log\frac{2k}{\delta}}{n}}\nonumber\\
\shortintertext{It is now assumed that all $\mathbb{E}[r_{i}]$ terms are estimated to at least $\epsilon$ accuracy. Substituting this back into the definition of the k-step TD update we get}\nonumber
\hat{Q}_{n+1}(s, a) \mathit{-} Q(s, a) = \frac{1}{n}\sum_{x=1}^{n}\bigg(r_{0} + \eta r_{1} + ... + \eta^{k-1}r_{k-1} + \eta^{k}Q_{n}(s_{k}, a_{k}) \bigg) \mathit{-} Q(s, a)\nonumber\\
= \sum_{i=0}^{k-1}\eta^{i} \bigg(\frac{1}{n}\sum_{i=1}^{n}r_{i}^{(x)} -\mathbb{E}[r_{i}] \bigg) + \eta^{k} \bigg(\frac{1}{n}\sum_{i=1}^{n} Q_{n}(s_{k}^{i}, a_{k}^{i}) -\mathbb{E}[Q(s_{k},a_{k})] \bigg)\\
\shortintertext{where in the second line we re-expressed the value in terms of a sum of k rewards. We now upper bounded the difference from $E[r_{i}]$ by $\epsilon$ to get}\nonumber
\hat{Q}_{n+1}(s, a) \mathit{-} Q(s, a) \leq \sum_{i=0}^{k-1}\eta^{l} \epsilon + \eta^{k} \bigg(\frac{1}{n}\sum_{i=1}^{n} Q_{n}(s_{k}^{i}, a_{k}^{i}) -\mathbb{E}[Q(s_{k},a_{k})] \bigg)\nonumber\\
\shortintertext{The variance term arises from the deviation of the empirical average of rewards from the true expected reward:}\nonumber
\epsilon \bigg( \frac{1-\eta^{k}}{1 - \eta} \bigg)\nonumber\\
\leq \epsilon \bigg( \frac{1-\eta^{k}}{1 - \eta} \bigg) + \eta^{k} \bigg(\frac{1}{n}\sum_{i=1}^{n} Q_{n}(s_{k}^{i}, a_{k}^{i}) -\mathbb{E}[Q(s_{k},a_{k})] \bigg)\\
\shortintertext{The bias term is propagated through bootstrapping:}\nonumber
\eta^{k}\Delta^{\hat{Q}_{\eta}}_{n-1}\nonumber\\
\shortintertext{and then the second term is bounded by $\Delta^{\hat{Q}_{\eta}}_{n-1}$ by assumption. Therefore, the combination of these terms results in the comprehensive bound for SARSA:}
\Delta^{\hat{Q}_{\eta}}_{n} \leq \epsilon \bigg( \frac{1-\eta^{k}}{1 - \eta} \bigg) + \eta^{k}\Delta^{\hat{Q}_{\eta}}_{n-1}
\end{align}
\endgroup
\end{proof}

\begin{proof}{Theorem \ref{thm:thm4}}
\begingroup
\allowdisplaybreaks
\begin{gather}\label{prf:thm4}\nonumber
\shortintertext{Phased $TD(\Delta)$ update rules for $m \geq 1$:}
\begin{split}
\hat{D}_{m,n}(s,a) = \frac{1}{n}\sum_{x-1}^{n}\biggl(\sum_{i=1}^{k_{m}-1} (\eta_{m}^{i} - \eta_{m-1}^{i})r_{i}^{(x)} + (\eta_{m}^{k_{m}} - \eta_{m-1}^{k_{m}})\hat{Q}_{\eta_{m-1}}(s_{k}^{(x)}, a_{k}^{(x)}) + \eta_{m}^{k_{m}} \hat{D}_{m}(s_{k}^{(x)}, a_{k}^{(x)})\biggr)
\end{split}
\shortintertext{According to the multi-step update rule \ref{MSTDSARSA} for $m \geq 1$:}
D_{m}(s_{n}, a_{n}) = \mathbb{E}\bigg[\sum_{i=1}^{k_{m}-1} \bigg(\eta_{m}^{i} - \eta_{m-1}^{i} \bigg)r_{i} + \bigg(\eta_{m}^{k_{m}} - \eta_{m-1}^{k_{m}} \bigg) Q_{\eta_{m-1}}(s_{k}, a_{k}) + \eta_{m}^{k_{m}} D_{m}(s_{k},a_{k}) \bigg]\\
\shortintertext{Then, subtracting the two expressions gives for $m \geq 1$:}\nonumber\\
\begin{split}
\hat{D}_{m,n}(s,a) - D_{m}(s_{n}, a_{n}) &= \sum_{i=1}^{k_{m}-1} \bigg(\eta_{m}^{i} - \eta_{m-1}^{i} \bigg)\bigg( \frac{1}{n}\sum_{x-1}^{n}r_{i}^{(x)} - \mathbb{E}[r_{i}]\bigg) + \bigg(\eta_{m}^{k_{m}} - \eta_{m-1}^{k_{m}} \bigg)\nonumber\\ & \bigg(\sum_{q=0}^{m-1} \hat{D}_{q}(s_{k}^{(x)},a_{k}^{(x)}) - \mathbb{E}[D_{m}(s_{k},a_{k})]  \bigg)
+ \eta_{m}^{k_{m}} \bigg(D_{m}(s_{k}^{(x)},a_{k}^{(x)}) - \mathbb{E}[D_{m}(s_{k},a_{k})] \bigg)
\end{split}
\shortintertext{Assume that $k_{0}\leq k_{1}\leq ... k_{m}=k$, the D estimates share at most $k_{m}=k$ reward terms $\frac{1}{n}\sum_{x-1}^{n}r_{i}^{(x)}$, Using Hoeffding inequality and union bound, it follows that, with probability $1 - \delta$, each k empirical average reward $\frac{1}{n}\sum_{x-1}^{n}r_{i}^{(x)}$ deviates from the true expected reward $\mathbb{E}[r_{i}]$ by at most $\epsilon = \sqrt{\frac{2log(2k/\delta)}{n}}$. Hence, with probability $1 - \delta$, $\forall m \geq 1$, the following holds:}\nonumber\\
\Delta_{n}^{m} \leq \epsilon \sum_{i=1}^{k_{m}-1} \bigg(\eta_{m}^{i} - \eta_{m-1}^{i} \bigg) + \bigg(\eta_{m}^{k_{m}} - \eta_{m-1}^{k_{m}} \bigg) \sum_{q=0}^{m-1} \Delta_{n-1}^{q} + \eta_{m}^{k_{m}} \Delta_{n-1}^{m}\nonumber\\
= \epsilon \bigg(\frac{1-\eta_{m}^{k_{m}}}{1 - \eta_{m}} - \frac{1 - \eta_{m-1}^{k_{m}}}{1 - \eta_{m-1}} \bigg) + \bigg(\eta_{m}^{k_{m}} - \eta_{m-1}^{k_{m}} \bigg) \sum_{q=0}^{m-1} \Delta_{n-1}^{q} + \eta_{m}^{k_{m}} \Delta_{n-1}^{m}
\shortintertext{ and $\Delta_{n}^{0} \leq \epsilon \frac{1 - \eta_{0}^{k_{0}}}{1 - \eta_{0}} + \eta_{0}^{k_{0}}\Delta_{n-1}^{0}$}
\shortintertext{Summing the two previous inequalities gives:}\nonumber
\sum_{m=0}^{M} \Delta_{n}^{m} \leq  \epsilon \frac{1 - \eta_{0}^{k_{0}}}{1 - \eta_{0}} + \epsilon \sum_{m=1}^{M} \bigg(\frac{1-\eta_{m}^{k_{m}}}{1 - \eta_{m}} - \frac{1 - \eta_{m-1}^{k_{m}}}{1 - \eta_{m-1}} \bigg) + \sum_{m=1}^{M} \bigg(\eta_{m}^{k_{m}} - \eta_{m-1}^{k_{m}} \bigg) \sum_{q=0}^{m-1} \Delta_{n-1}^{q} + \eta_{m}^{k_{m}} \Delta_{n-1}^{m}\\
= \underbrace{\epsilon \frac{1-\eta_{M}^{k_{M}}}{1 - \eta_{M}} + \epsilon \sum_{m=0}^{M-1}  \frac{\eta_{m}^{k_{m+1}}-\eta_{m}^{k_{m}}}{1 - \eta_{m}}}_{\text{(*)variance term}}   + \underbrace{\sum_{m=1}^{M} \bigg(\eta_{m}^{k_{m}} - \eta_{m-1}^{k_{m}} \bigg) \sum_{q=0}^{m-1} \Delta_{n-1}^{q} + \eta_{m}^{k_{m}} \Delta_{n-1}^{m}}_{\text{(**)bias term}}\\
\shortintertext{Let’s focus now further on the bias term (**)}\nonumber\\
\sum_{m=1}^{M} \bigg(\eta_{m}^{k_{m}} - \eta_{m-1}^{k_{m}} \bigg) \sum_{q=0}^{m-1} \Delta_{n-1}^{q} + \eta_{m}^{k_{m}} \Delta_{n-1}^{m} = \sum_{q=0}^{M-1} \sum_{m=q+1}^{M} \bigg(\eta_{m}^{k_{m}} - \eta_{m-1}^{k_{m}} \bigg) \Delta_{n-1}^{q} + \sum_{m=1}^{M} \eta_{m}^{k_{m}} \Delta_{n-1}^{m}\nonumber\\
= \sum_{q=0}^{M-1}\Delta_{n-1}^{q}\bigg(\sum_{m=q+1}^{M} \eta_{m}^{k_{m}} - \sum_{m=q}^{M-1} \eta_{m}^{k_{m+1}} \bigg)+ \sum_{m=1}^{M} \eta_{m}^{k_{m}} \Delta_{n-1}^{m} \nonumber\\
= \sum_{q=0}^{M-1}\Delta_{n-1}^{q}\bigg(\sum_{m=q+1}^{M-1} (\eta_{m}^{k_{m}} - \eta_{m}^{k_{m+1}}) + (\eta_{M}^{k_{m}} - \eta_{q}^{k_{q+1}})  \bigg)+ \sum_{m=1}^{M} \eta_{m}^{k_{m}} \Delta_{n-1}^{m} \nonumber\\
= \sum_{q=0}^{M-1}\sum_{m=q+1}^{M-1}(\eta_{m}^{k_{m}} - \eta_{m}^{k_{m+1}})\Delta_{n-1}^{q} + \eta_{M}^{k_{m}} \sum_{m=0}^{M} \Delta_{n-1}^{m} + \sum_{m=0}^{M-1} (\eta_{m}^{k_{m}} - \eta_{m}^{k_{m+1}}) \Delta_{n-1}^{m} \nonumber\\
= \sum_{q=0}^{M-1}\sum_{m=q}^{M-1}(\eta_{m}^{k_{m}} - \eta_{m}^{k_{m+1}})\Delta_{n-1}^{q} + \eta_{M}^{k_{m}}\sum_{m=0}^{M} \Delta_{n-1}^{m} \nonumber\\
= \sum_{m=0}^{M-1}(\eta_{m}^{k_{m}} - \eta_{m}^{k_{m+1}})\sum_{q=0}^{m}\Delta_{n-1}^{q} + \eta_{M}^{k_{m}}\sum_{m=0}^{M} \Delta_{n-1}^{m}\\
\shortintertext{Finally, the following is obtained:}\nonumber\\
\sum_{m=0}^{M} \Delta_{n}^{m} \leq \underbrace{\epsilon \bigg( \frac{1-\eta^{k}}{1 - \eta} \bigg)+ \epsilon \bigg(\sum_{m=0}^{M-1} \frac{\eta_{m}^{k_{m+1}}-\eta^{k_{m}}_{m}}{1-\eta_{m}} \bigg)}_{\text{variance reduction}}+ \underbrace{\sum_{m=0}^{M-1}\bigg( \eta^{k_{m}}_{m} - \eta^{k_{m+1}}_{m} \bigg)\sum_{q=0}^{m}\Delta_{n-1}^{q} + \eta^{k}\sum_{m=0}^{M}\Delta_{n-1}^{m}}_{\text{bias introduction}}
\end{gather}
\endgroup
\end{proof}

\section[\appendixname~\thesection]{Hyperparameters}
\label{appendix-hyper}
The experiments primarily utilize the hyperparameters recommended by Romoff 771
et al.\cite{romoff2019separating}, Schulman et al.\cite{schulman2017proximal} and Kostrikov\cite{pytorchrl}. The experiments utilize the following randomly generated seeds: 125125, 513, 90135, 81212, 3523401, 15709, 17, 0, 8412, and 1153780. Additional hyperparameter configurations are detailed in Table \ref{hyperparameter-table}.
\begin{table}[H]
\centering
\caption{Hyperparameter configurations and their corresponding values.}
\label{hyperparameter-table}
\begin{tabular}{l|r|}
\cline{1-2}
\multicolumn{1}{|l|} {Hyperparameter} & Value \\
\hline
\multicolumn{1}{|l|} {seed (10)} & np.random.randint(40, 45) \\
\hline
\multicolumn{1}{|l|} {Learning Rate ($\alpha$)} & [0.1, 0.3, 0.5, 0.7, 1.0] \\
\hline
\multicolumn{1}{|l|} {Discount Factor ($\eta$)} & [0.992, 0.95, 0.996, 0.99, 1.0] \\
\hline
\multicolumn{1}{|l|} {k-Step} & [1, 4, 8, 16, 32, 128, 256] \\
\hline
\multicolumn{1}{|l|} {Number of Episodes} & 100000 \\
\hline
\multicolumn{1}{|l|} {Max Epsilon} &  0.4\\
\hline
\multicolumn{1}{|l|} {Clipping parameter} &  0.1\\
\hline
\multicolumn{1}{|l|} {Number of actors} & 8\\
\hline
\multicolumn{1}{|l|} {Horizon(T)} & 129\\
\hline
\multicolumn{1}{|l|} {Number of Epochs} & 3\\ 
\hline
\multicolumn{1}{|l|} {Lambda ($\lambda$)} & 0.95\\ 
\hline
\end{tabular}
\end{table}

\end{document}